\documentclass[twocolumn,letterpaper]{IEEEAerospaceCLS}  % only supports two-column, letterpaper format

\usepackage[]{graphicx}    % We use this package in this document
\usepackage{amsmath}
\usepackage{float}
\usepackage{hyperref}
\usepackage{subfig}
\usepackage{stackrel}
\usepackage{enumitem}

\usepackage{todonotes}
\newcommand{\ignore}[1]{}  % {} empty inside = %% comment
\usepackage{import}
\newcommand{\slugs}[0]{\texttt{slugs} }

\newcommand{\inquote}[1]{``#1''}

\begin{document}
\title{Resolving Ambiguity via Dialogue to Correct Unsynthesizable Controllers for Free-Flying Robots}

\author{%
Joshua Rosser, Jacob Arkin, Siddharth Patki, Thomas M. Howard\\ 
Department of Electrical Computer Engineering \\
University of Rochester\\
Rochester, NY 14627\\
jrosser2@ur.rochester.edu, j.arkin@ur.rochester.edu, spatki@ur.rochester.edu, thoward@ece.rochester.edu
%%%% IMPORTANT: Use the correct copyright information--IEEE, Crown, or U.S. government. %%%%%
\thanks{\footnotesize 978-1-6654-9032-0/23/$\$31.00$ \copyright2023 IEEE}              % This creates the copyright info that is the correct 2023 data.
%\thanks{{U.S. Government work not protected by U.S. copyright}}         % Use this copyright notice only if you are employed by the U.S. Government.
%\thanks{{978-1-6654-9032-0/23/$\$31.00$ \copyright2023 Crown}}          % Use this copyright notice only if you are employed by a crown government (e.g., Canada, UK, Australia).
%\thanks{{978-1-6654-9032-0/23/$\$31.00$ \copyright2023 European Union}}    % Use this copyright notice is you are employed by the European Union.
}

\maketitle

\thispagestyle{plain}
\pagestyle{plain}

\maketitle

\thispagestyle{plain}
\pagestyle{plain}

\begin{abstract}
% An abstract of 250 to 500 words should concisely describe the work being reported, its methodology, principal results and significance.
For human-robot teams that operate in space, safety and robustness are paramount.
In situations such as habitat construction, station inspection, or cooperative exploration, incorrect assumptions about the environment or task across the team could lead to mission failure.
Thus it is important to resolve any ambiguity about the mission between teammates before embarking on a commanded task.
The safeguards guaranteed by formal methods can be used to synthesize correct-by-construction reactive controllers for a robot using Linear Temporal Logic.
If a robot fails to synthesize a controller given an instruction, it is clear that there exists a logical inconsistency in the environmental assumptions and/or described interactions.
These specifications however are typically crafted in a language unique to the verification framework, requiring the human collaborator to be fluent in the software tool used to construct it.
Furthermore, if the controller fails to synthesize, it may prove difficult to easily repair the specification.
Language is a natural medium to generate these specifications using modern symbol grounding techniques.
Using language empowers non-expert humans to describe tasks to robot teammates while retaining the benefits of formal verification.
Additionally, dialogue could be used to inform robots about the environment and/or resolve any ambiguities before mission execution.
This paper introduces an architecture for natural language interaction using a symbolic representation that informs the construction of a specification in Linear Temporal Logic.
The novel aspect of this approach is that it provides a mechanism for resolving synthesis failure by hypothesizing corrections to the specification that are verified through human-robot dialogue.
Experiments involving the proposed architecture are demonstrated using a simulation of an Astrobee robot navigating in the International Space Station.
\end{abstract}

\tableofcontents

%%%%%%%%%%%%%%%%%%%%%%%%%%%%%%%%%%%%%%

\section{Introduction}
%It is clear that when dealing in a space setting, a heterogenous combination of the aforementioned methods would provide the most robust system \cite{Webster2020}, but in this work we choose to focus on the formal method of verification by synthesis.
%
%A benefit of using a LTL as a specification language is that failure to synthesize can be leveraged to work to resolve the inconsistencies between what the robot is being asked to complete, and what the human believes the robot can complete.
%
%In \cite{rosser2020a} it was demonstrated that by providing declarative knowledge to the Astrobee, the specification could be repaired enabling the robot to complete its navigation task, but expecting the human to know ahead of time what is missing from the robot's understanding is not conducive.
%
%By leveraging the nature of the DCGs and the symbol space, we can rollout a set of hypothetical worlds under which the specification would be synthesizable.
%
%The missing LTL propositions can then be used via inverse semantics to ask a targeted question to the human to in order to resolve its confusion.
%
%Upon verification from the human, the robot can update its world model to synthesize the verifiably correct controller to complete the task.
%
%We demonstrate the pipeline in the simulated ISS in a capsule navigation task with an Astrobee robot.
%
%Because slugs is so fast, each rollout can be checked to see if its contribution is verifiable.

 %%%%%%%%%%%%%%%%%%%%%%%%%%%%%%%%%%%%%%%%%%%%%%%%%%%
There exists a long and rich history of using robotics to expand our understanding of the solar system.
%
%At a time where our interests are pointed towards the likes of putting humans back on Moon and to Mars, coexisting during a time of a great expanse in robot capabilities, a natural partnership between human and robot emerges.
%
Current missions of robotic exploration have humans remotely guide robot assets to perform a variety of science operations \cite{Goldberg2002}.
Future missions may require humans and robots to cooperatively perform tasks in-situ.
There have been many studies involving astronaut-robot teams \cite{Fong,Pedersen2006,trevino2000first,Sierhuis2005}.
Two examples of mission-oriented experiments involving astronaut-rover teams include planetary outpost assembly \cite{Medina2011} and base camp operations \cite{Diftler2007}.
Like the methods explored in this paper, the Human-Robot Interaction Operating System (HRI/OS) emphasized natural language and dialogue as a medium to share information amongst agents in space relevant contexts \cite{Fong2006}.
Modern approaches to natural language processing (NLP), specifically those that involve grounded language communication \cite{Howard2022,howard14a,paul18a,tellex2011a}, have enabled humans and robots to perform tasks with a mutual understanding of their surrounding world.

\begin{figure}[H]
    \centering
    \includegraphics[width=1.0\columnwidth]{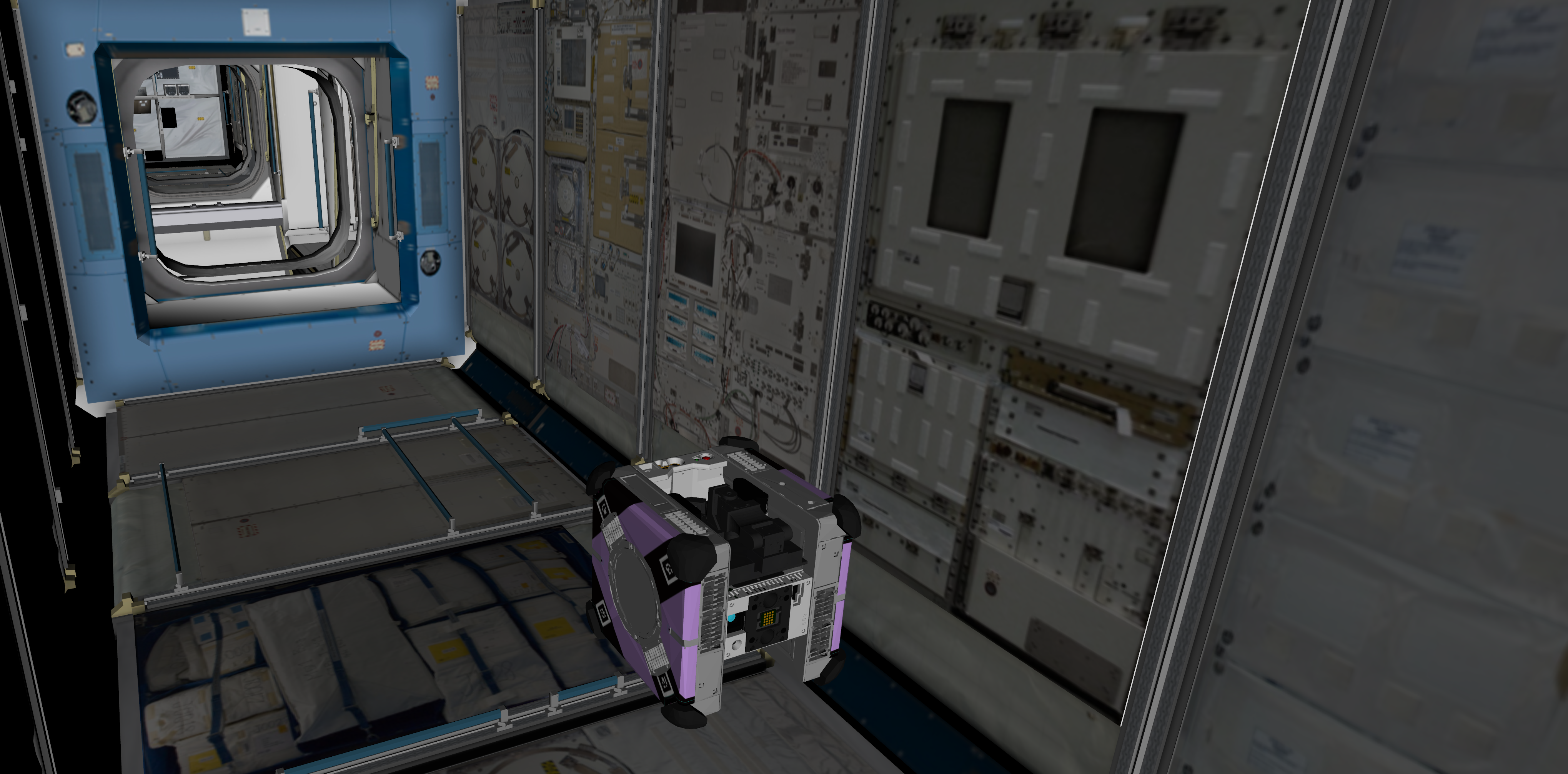}
    \caption{Simulation of Astrobee in the JEM/Kibo capsule of the International Space Station}
    \label{fig:astrobee}
\end{figure}

Formal methods, which are various mathematical languages, techniques, and tools that can specify and verify the design specifications of a controller \cite{clarke1996a}, have become more pervasive in robotics as the consequences of incorrect actions have become more severe.
Specifications that place guarantees on robot behavior given a set of assumptions about the world in which the robot operates can be used to generate controllers that drive robots actions.
This generation process is referred as synthesis and is possible by utilizing techniques dictated with formal methods.
One such logic that can be used as a language to write a specification is the atomic propositions of Linear Temporal Logic (LTL) \cite{Manna1991TheTL,Pnueli1977TheTL}.
This paper studies the problem of repairing specifications interpreted from natural language that are unable to synthesize controllers through dialogue.  
By combining techniques from grounded language communication and formal verification, a system is proposed that enables a robot to synthesize a correct-by-construction controller from an instruction provided in the context of an incomplete environment model.
We demonstrate this pipeline in a hypothetical human-robot teaming scenario with a simulation of the Astrobee robot \cite{smith16a} in the International Space Station (ISS) as seen in Figure \ref{fig:astrobee} that is commanded to navigate between a variety of space station modules.
% END WITH THIS

%It is clear that enabling a human and robot to engage in dialogue with one another is a straightforward means to collaboration, but it raises the issue of how to prevent robots from acting unpredictably in these extreme environments.
%In space, where time is the most valuable resource, even the smallest mistake driven by a misunderstanding between the mental models of the human and robot can be costly, upwards of catastrophic.
%Consider the case where a rover is sent to the far side of the Moon where communications may cut out, the human would want to be sure that the robot will successfully complete its task prior to embarking on the journey.
%It is clear that guaranteed safety and predictability among human-robot teams is necessary when operating in a space environment whether astronauts working in situ on a planetary surface, or in standardizing the New Space sector in situations such as space tourism and mining \cite{cardoso2021verification}. %
%

\section{Related Work}
The aerospace community has an extensive record of using verifiable methods to evaluate the behavior of autonomous systems \cite{Gao2016,cardoso2021verification}.
%
%\cite{cardoso2021verification} argues for why it is a topic of ever increasing interest especially if autonomy is to become more prevalent in these systems.
%
A variety of verification methods are outlined in \cite{kress2021formalizing} as possible methods to bound robot behavior when operating with and around humans.
\textit{Synthesis} involves leveraging the properties of either discrete or probabilistic temporal logics to generate a finite state machine from a specification \cite{kress2018a}. 
In \textit{formal verification} behavior dictated by a specification is guaranteed by extensively probing the system via model checking \cite{Clarke2018} or via checking the specification against theoretical axioms \cite{Hoare1969}.
With \textit{runtime monitoring} a system is observed with various monitors at runtime to check if at any point the specification is violated by the resultant behavior \cite{LEUCKER2009293}.
Finally, \textit{test-based} verification can be from either simulation \cite{AraizaIllan2015} or end-user experiments \cite{Salem2015} running the system through iterations of realistic scenarios gaining a measure of predictability of the system.
Despite simulated testing requiring the fewest abstractions from the reality the system will face, it may prove to be the most time intensive to complete.
Correct-by-construction reactive controllers synthesized from specifications constructed using LTL have been previously used in mission and motion planning tasks \cite{kress2009a}.
Full LTL remains computationally intractable \cite{pnueli1989a} for a synthesis process, therefore a subset of LTL can be used to construct a specification.
Consistent with the work in \cite{rosser2020a}, we adopt the use of the GR(1) fragment \cite{BLOEM2012911} utilized in the synthesis toolkit \slugs \cite{ehlers16a}.
\slugs synthesizes verifiably correct controllers that guarantee system behavior given a set of environmental assumptions in singly-exponential time complexity \cite{clarke1996a,ehlers15a}.
Clearly, there are benefits to employing a synthesis method such as \slugs, but there still remains the problem of how a specification is constructed.
In most toolkits, as in \slugs, crafting the actual specification requires expert knowledge of the corresponding scripting language.
A number of works have tackled the problem of mapping natural language to LTL \cite{kress2008a,finucane2010a,Lignos2014a}.
One such approached used probabilistic graphical models to write LTL specifications from natural language in a tabletop environment \cite{boteanu16a}.

%For example, as part of the Mobile Agents Architecture at NASA Ames\cite{Sierhuis2005}, a multiagent workflow language called Brahms \cite{SierhuisSocialScienceandInformatics} was used to model the behavior between the astronaut-rover team.
%
%Brahms is a toolkit that has its own modeling language to generate specifications that outline an agents beliefs, activities, and movements, however, to verify the guarantees of the decision making process on the agent level, the Brahms specification has to be translated into another model checker language consistent with Promela 2 \todo{needs citation, should we even have this}.
%
%If an astronaut had to make an adjustment in situ in realtime, but still wanted to leverage the same guarantees of a formal method, this is not conducive.
%

%
Another benefit to synthesizing a correct-by-construction controller is that synthesis failure can be used as an indicator for a misunderstanding between the human and robot.
Using grounded language to repair LTL specifications was demonstrated in previous work \cite{boteanu17a}.
In this work a query related to the assumptions about the environment was generated for the human via templated language.
The approach described in \cite{rosser2020a} addressed the problem of providing declarative knowledge \cite{arkin2020} to update the robot's world model and prevent synthesis failure of subsequent instructions.
It is however impractical to assume that humans will always anticipate every gap in the specification before an instruction is given.
A more robust framework is one that enables the robot to accept declarative knowledge and ask questions to resolve ambiguities about the interpreted task that prevent controller synthesis.
This paper describes a method for resolving such ambiguities by testing hypothetical worlds that correspond to realizable specifications and confirming such worlds through human-robot dialogue.
This work differs from \cite{boteanu17a} in that language queries provided by the robot that correspond to specification updates is generated from a learned model of language.
Additionally the techniques described in \cite{Williams2013} also introduce an algorithm to resolve spatial reference resolutions and add information to its world model through language with use of the \texttt{SPEX} pipeline.
This pipeline, however, does not leverage a verifiable controller or any type of formal method.
%

%
%By then checking if a given hypothetical world repairs the specification by rerunning synthesis, the initial language symbol can be used in the inverted language model to ask a targeted question seeking to validate the proposed logic.
%
%Upon human verification the robot uses the synthesized controller to complete the outlined task.
%
%
%However, a main difference in our work is the ability to provide knowledge to the robot in the form of declarative statements and of the robot to query the human with a targeted question to get help when confronted with processing a region with an unknown spatial relation and a dialogue is not being driven through the use of formal methods.
%

%

\section{Technical Approach}
\subsection{Natural Language Understanding}
One formulation of grounded natural language understanding for robots is a discriminative optimization problem in which the goal is to find the most likely correspondence $\Phi^{*}$ between a known set of semantic symbols $\Gamma$ and a known utterance $\Lambda$ given the context of the surrounding environment $\Upsilon$:

\begin{equation} \label{eq:high-level-problem-formulation}                       
\Phi^{*} = \underset{\Phi \in \mathbf{\Phi}}{\arg\max} \hspace{1mm} p( \Phi \vert \Gamma, \Lambda, \Upsilon )
\end{equation} 
This is challenging to compute directly for a variety of reasons, such as the diversity of both language and the associated concepts, and the complexity of realistic environments, among other reasons.
A modern class of approaches use factor graphs to make this problem tractable \cite{tellex2011a,howard14a,paul18a}.
The approach used in this work, Distributed Correspondence Graphs (DCGs) \cite{howard14a}, factorizes over both constituents of language $\Lambda = \{ \lambda_1, \lambda_2, \dots \}$ and symbolic constituents of semantic concepts $\Gamma = \{ \gamma_1, \gamma_2, \dots \}$.
In particular, we represent language as constituency parse trees, a syntactic representation with phrase constituents.
DCGs make use of the compositionality of language by assuming that 1) the semantics of individual sibling phrases in the parse tree are conditionally independent and 2) the semantics of the full utterance is composed from the tree structure.
In order to linearize the exponentially-sized set of possible correspondences $\Phi_i \in \Phi$ between an individual phrase $\lambda_i \in \Lambda$ and the set of semantic concepts $\Gamma$, DCGs further assume conditional independence of correspondence of individual constituents of the semantic symbols.
Therefore, each factor in the factor graph computes the likelihood of the correspondence $\phi_{ij}$ between the $i^{th}$ phrase $\lambda_i$ and the $j^{th}$ semantic symbol $\gamma_j$, resulting in the following reformulation:
\begin{equation} \label{eq:dcg-nlu}                                              
\Phi^{*} =  \underset{\phi \in \mathbf{\Phi}}{\arg\max}                          
\prod\limits_{i=1}^{\lvert \Lambda \rvert}                                       
\prod\limits_{j=1}^{\lvert \Gamma \rvert}                                        
p( \phi_{ij} \vert \gamma_{j}, \Phi_{C_{i}}, \lambda_{i}, \Upsilon )             
\end{equation}  
Rather than compute each factor directly, we use a log-linear model consisting of expert-designed feature indicator functions and aligned weights optimized with respect to a fully supervised training corpus.
For more details about DCGs, we refer to \cite{paul18a}.
The choice of symbolic semantic representation is fundamental to the capability of DCGs since it determines the scope of concepts expressed in language that can be understood.
In this work, we use a semantic representation consisting of four main categories:

\begin{enumerate}[leftmargin=0.75cm]
	\item Objects: entities (e.g.  a laptop) or regions (e.g. the Harmony capsule) in the world, typically populated a priori or via a perception system
	\item Connectivity Relations: connections between regions in the world
	\item Spatial Relations: spatial relationships between entities in the world
	\item Actions: actions for the robot to execute (e.g. navigation, object retrieval, room inspection)
\end{enumerate}

\subsection{Synthesis of Correct-by-Construction Reactive Controllers}

Correct-by-construction controllers are guaranteed to operate within the assumptions outlined in a specification used to generate the controller.
A variety of logics can be used to craft specifications.
In this work LTL is used to describe both the behavior of the robot and the environment while also utilizing its temporal operators at the concession that guarantees are made over an infinite time horizon.
As described in \cite{kress2018a} LTL formulae can be constructed by the given grammar \cite{Manna1991TheTL} if $\pi\in AP$ is assumed of which $AP$ as the set of atomic propositions in which $\pi$ is a Boolean variable.
The grammar is provided by Equation~\ref{eq:ltl} where $\neg,\; \lor$ are Boolean operators negation'' and ``disjunction'' and $\bigcirc,\; \mathcal{U}$ are temporal operators ``next'' and ``until''.
\begin{equation}\label{eq:ltl}
\varphi::=\pi \: | \: \neg \varphi \: | \: \varphi \lor \varphi \: | \: \bigcirc \varphi \: | \: \varphi \mathcal{U} \varphi \:
\end{equation}
From there \inquote{conjunction}: $\varphi \land \varphi$, \inquote{implication}: $\varphi \Rightarrow \phi$, and \inquote{equivalence}: $\varphi \Leftrightarrow \phi$ can be defined as well as the additional temporal operators \inquote{eventually}: $\Diamond \varphi=\top \mathcal{U}\varphi$ and \inquote{always}: $\Box \varphi = \neg \Diamond \neg \varphi$.
Consistent with the work in \cite{rosser2020a}, reactive controllers can be synthesized in realtime using the \slugs toolkit \cite{ehlers16a} by incorporating the modifications on the full LTL grammar made with GR(1) fragment \cite{BLOEM2012911}.
How the GR(1) fragment modifies LTL is described in \cite{kress2018a}.
Following the notation from that paper, consider the set of atomic propositions $AP= \mathcal{X} \cup \mathcal{Y}$ where $\mathcal{X}$ are the input propositions and $\mathcal{Y}$ are the output propositions that compose a \slugs specification.
$\mathcal{X}$ are Boolean representations of the environment as observed by the sensors on Astrobee, and $\mathcal{Y}$ are Boolean representations of the individual actions (i.e. moving) available to the robot.
For our experiments the capabilities of the sensors are limited to detecting if the Astrobee resides in a particular capsule of the simulated ISS, and the only permissible action is moving between the ISS capsules. 
The fragment consists of LTL formulae that satisfy $\varphi=\left(\varphi_{e} \Rightarrow \varphi_{s} \right)$, where $\varphi_{e}$ is an assumption about the environment in LTL, and $\varphi_{s}$ define the guarantees on how the robot will act given these assumptions.
The following define $\varphi_{e}$ and $\varphi_{s}$:
\begin{equation}
\varphi_{e}=\varphi^{e}_{i} \land \varphi^{e}_{t} \land \varphi^{e}_{l},\;\;\;\;
\varphi_{s}=\varphi^{s}_{i} \land \varphi^{s}_{t} \land \varphi^{s}_{l}
\end{equation}

Thereby, the three components of the GR(1) fragment (initial conditions, transitions, and liveness), as similarly explained in~\cite{kress2018a}, can be described as follows:
\begin{enumerate}[leftmargin=1.25cm,label={[C\arabic*]}]
    \item The initial conditions of the propositions under $\mathcal{X}$ and $\mathcal{Y}$ at the start of the reactive controller are set by the Boolean formulae $\varphi_{i}^{e}$ and $\varphi_{i}^{s}$.
    \item The transitional assumptions of the environment and guarantees of the robot are defined by $\varphi_{t}^{e}$ and $\varphi_{t}^{s}$. The operator $\Box$ for always is applied to both formulae, providing a guarantee to be True at any state of the controller.
    \item The environment and system liveness, $\varphi_{l}^{e}$ and $\varphi_{l}^{s}$, are formulae describing the goals of each respective player. The operator $\Diamond$ for eventually is applied to both formulae, providing a guarantee to be satisfied at some point in the future.
\end{enumerate}
To synthesize a controller using the GR(1) fragment, \slugs solves a $\mu$-calculus fixpoint equation on a two-player game structure where the environment player and system player alternate making permissible moves as defined by the transition formulae with the environment always making the first move.
The environment and system work to satisfy each respective liveness condition infinitely often.
A specification is said to be \textit{unrealizable} when there exists at least one environment where the system player cannot accomplish its goal(s), this will fail to synthesize a controller \cite{raman2013b}.
There are two modifications to the synthesis process adopted in this work. 
The first ensures that robot behavior is guaranteed no matter the permissible initial state of the state machine \cite{kress2009a}.
The second requires that the robot player will not seek behaviors that sabotage the environment player from reaching its goal \cite{ehlers15a}.
For more information on the GR(1) fragment and synthesis using \slugs please see \cite{Raman-RSS-13,ehlers15a,ehlers16a,kress2018a,raman2013b}.

\subsection{Reasoning Over Hypothetical Worlds}
This work introduces the concept of hypothetical worlds to the pipeline.
When synthesis of the controller fails, this is indicative of a misalignment between the models of the world held by human and the robot.
To mend this gap, the robot must reason over the differences by hypothesizing over potential worlds rooted in the current world representation, this can be accomplished by the robot asking targeted questions about its environment to the human.
The synthesis process is used to identify hypothetical worlds that repair the logical gap by dismissing those that do not synthesize a controller.
%
%If a hypothetical specification is realizable the human can then be queried about the theorized information as to whether it is valid to the human's reality.
%

%
The set of semantic concepts represented by the language model places constraints on the set of hypothetical worlds considered during the LTL controller synthesis search process.
Since the system generates a grounded language query pertaining to the difference in the actual world and the hypothetical world used by the successfully synthesized specification, the language model's set of semantic concepts must be sufficiently granular to represent that difference.
In order to enforce this constraint, we therefore rely on the instantiated DCG to provide the set of world-relevant concepts over which the hypothetical synthesis process will search.
For the particular set of scenarios demonstrated in this paper, these world changes are limited to the set of possible Connectivity Relations instantiated by the DCG according to the world model at the time of the initial utterance.
It is therefore possible to iterate over the set of possible Connectivity Relations, and at each Connectivity Relation append the contained information to a copy of the current world model, which is analyzed to generate a hypothetical specification.
If the addition of this information results in a realizable controller, that information can be used to query the human about its validity.
When information from a singular Connected Relation is already represented in the world model, it is merely passed over as it will result in an unrealizable specification.
If a hypothetical world is deemed to be consistent with the human's mental model, this new information augments the robot's world model.
To determine if a hypothetical world is consistent with human's mental model, the associated Connectivity Relation can be used to generate a query in the form of a question in natural language.

\subsection{Natural Language Generation for Issuing Queries}

In order to ask the human to confirm whether the hypothesized world state reflects the actual world, our system generates a template query populated by grounded language corresponding to the hypothesized world difference.
By construction, this hypothetical difference directly maps to the symbolic semantic concepts used by the language model.
Prior work has demonstrated the capacity for DCGs to be used for natural language generation by inverting the inference process \cite{Howard2022}.
If there is a known set of $\mathrm{True}$ and $\mathrm{False}$ corresponding semantic symbols (e.g., $\mathrm{True}$ for the hypothetical world difference, in this case),  natural language generation is the process of finding the most likely language $\Lambda^{*}$ in the context of the current world $\Upsilon$.
Formally, given a set of possible language $\mathbf{\Lambda} = \{ \Lambda_1, \Lambda_2, \dots \}$, a known set of correspondences $\hat{\Phi} = \{\hat{\phi}_1, \hat{\phi}_2, \dots \}$, the set of semantic symbols $\Gamma$, and a world model $\Upsilon$, the goal is to find the language $\Lambda^{*}$ that maximizes the likelihood:
\begin{equation} \label{eq:dcg-nlg}                                              
\Lambda^{*} =  \underset{\Lambda \in \mathbf{\Lambda}}{\arg\max}\hspace{1mm}                
p( \Phi = \hat{\Phi} \vert \Gamma, \Lambda, \Upsilon )                           
\end{equation}
For a constituency parse tree representation of language,  $\mathbf{\Lambda}$ consists of phrases and can be generated in multiple ways, such as directly from an existing corpus of examples or using a grammar model.
We approximate search for the most likely language as a sequence of NLU search processes.
For each phrase $\Lambda_i \in \mathbf{\Lambda}$, we construct a DCG and find the most likely correspondences per Equation \ref{eq:dcg-nlu}.
Once completed, we select $\Lambda_{i}^{*}$ as the phrase for which the most likely correspondences $\Phi_{i}^{*}$ matches the desired known correspondences $\hat{\Phi}$ with the highest likelihood.
This grounded corresponding phrase is used to populate a query template that is then sent to the human user.
The human is expected to reply either ``yes" or ``no", and the system responds accordingly.
If ``no", then the hypothetical world is not reality and the hypothetical synthesis search process continues; if ``yes", then the hypothetical world is actual, and the system executes the associated synthesizable state machine.

\subsection{System Architecture}

\begin{figure*}[h]
    \centering
    \includegraphics[width=1.0\textwidth]{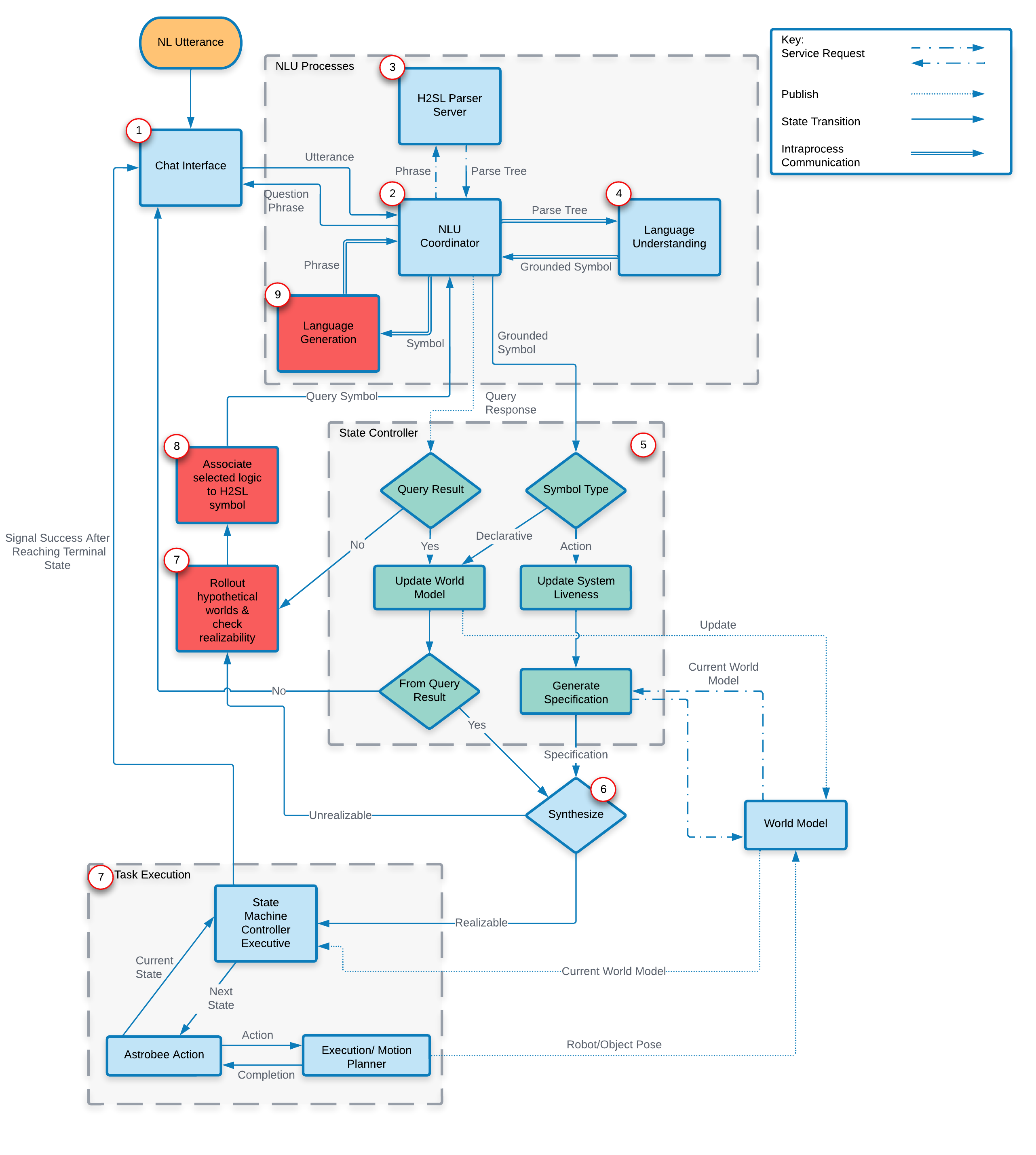}
    \caption{The proposed architecture for repairing specifications interpreted from natural language that are unable to synthesize controllers through dialogue.}
    \label{fig:system_dia}
\end{figure*}
Here we will discuss the system architecture representative of our pipeline refering to Figure~\ref{fig:system_dia}.
The baseline system architecture was introduced in \cite{rosser2020a}, namely the Chat Interface node (1), NLU Processes, State Controller (5), Synthesis process (6), Task Execution (7), and World Model node.
There are a few notable additions to the architecture including the H2SL Parser Server (3), Query Result process in (5), the process responsible for generating the set of hypothetical worlds (7), the process that associates the LTL formula to a semantic symbol (8), and the Language Generation process (9).
The processes in red (7, 8, 9) are all fundamental to generating queries to ask the human upon synthesis failure.
In addition, the world model node now updates the state machine controller to track the movement of the robot from capsule to capsule to inform the sensors in the state machine.
When an utterance in natural language is provided via text input to (1), a raw string is passed to (2), which then parses the utterance into a parse tree which is passed back to (2) and then to (4) which grounds the parse to the most likely corresponding semantic symbols.
When a natural language utterance is provided by a text input to (1), a raw string is passed to (2) which requests and then receives a parse tree from (3). 
The parse tree is sent to (4) which uses DCG to infer the most likely $\textrm{True}$ corresponding semantic symbols; in practice (2) enforces that there is exactly one $\textrm{True}$ corresponding symbol in the set.
That symbol is passed to (5) where it is deciphered as either declarative knowledge or an action.
If the symbol is declarative the contained information is used to update the world model.
The system then checks to see if the world model update was the result of declarative knowledge being provided or a query, if no as in the case that declarative knowledge is provided to the system prior to a command being given, the human is indicated via the chat interface that declarative knowledge was successfully received and processed.

If the symbol is an action symbol a LTL formula representing the system liveness $\varphi^{s}_{l}$ in the specification is generated and synthesis is attempted in (8).
The \slugs specification is written in the same way as with \cite{rosser2020a}, the LTL formulae describing the permissible environmental $\varphi^e_t$ and system $\varphi^s_t$ transitions in the \slugs specification are generated from information stored in a Kripke structure.
The Kripke structure encapsulates which state transitions are permissible.
If the specification is unrealizable and a controller fails to synthesize the process responsible for rolling out the set of hypothetical worlds is triggered.
For this process the Connectivity Relations are stored in a queue, when considering a hypothetical world a Connectivity Relation is popped off the queue and the encapsulated information which is of the type declarative knowledge is processed and appended to copy of the current world model.
A specification is generated consistent with the description above using this hypothetical world model, and the realizability is checked.
If unrealizable the system iterates on to the next Connectivity Relation in the queue and the process repeats until a realizable one is found.
If realizable the Connectivity Relation is passed along to (2) which triggers the language generation process in (8) that populates a query with an associated grounded utterance such as \inquote{is the kibo capsule connected to the columbus capsule?} via the process outlined in \textit{Natural Language Generation for Issuing Queries}.
This query is then provided to the human in (1), and the system waits for a response of yes or no validating or invalidating the query.
This result is passed to the State Controller Query Result process by means of (2).
If the response is \inquote{no}, then (7) is notified and considers the next hypothetical world generated from the next Connectivity Relation in the queue.
If the response is \inquote{yes}, then the hypothetical world in question is accepted as the current world model and synthesis ensues.
Now with a realizable specification, the verifiably correct reactive controller is generated and is passed to the Task Execution processes (7).
The robot begins its action sequence as dictated by the state machine until exiting upon reaching the goal state.

\section{Experimental Design}
All experiments take place in a simulated ISS with a singular simulated Astrobee robot.
The simulator is provided as a part of the Astrobee Flight Software hosted on \href{https://github.com/nasa/astrobee}{https://github.com/nasa/astrobee}.
For our experiments we command two navigation tasks between three of the six modules of the ISS: JEM/Kibo, Harmony, and Columbus.
In our reported figures from the simulator the order of the capsules is always as follows: Columbus is leftmost and connected to Harmony in the middle which is connected to JEM/Kibo rightmost.
State-estimation, planning, and all corresponding executive function is handled by functionality from the Astrobee software package upon receiving waypoints to travel to from the State Machine Controller Executive in Figure~\ref{fig:system_dia} as dictated by the synthesized reactive controller.
As far as the synthesized state machine is concerned the Astrobee is considered \inquote{in} a certain module when it is within 2.0 m of the centroid of the convex hull of the capsule.
The physical Astrobee system consists of three cube-shaped free-flying robots equipped with various perception sensors, a manipulator, and other peripherals for interacting with astronauts and partaking in onboard activities \cite{smith16a}.
We execute three different scenarios to demonstrate the functionality of the pipeline, in each case the robot has no knowledge of the connectivity of any of the capsules merely that the capsules Kibo/JEM, Harmony, and Columbus exist.

\begin{enumerate}[leftmargin=1.25cm,label={[E\arabic*]}]
    \item In this case the Astrobee unit starts off at the centroid of the Columbus capsule and is provided two declarative knowledge statements \inquote{the kibo capsule is connected to the harmony capsule} followed by \inquote{the harmony capsule is connected to the columbus capsule}. The command \inquote{go to the kibo capsule} is then given.

    \item In this case the Astrobee unit starts off at the centroid of the Kibo capsule and is given an single declarative knowledge statement \inquote{the kibo capsule is connected to the harmony capsule}, followed by the command \inquote{go to the columbus capsule}.

    \item As in experiment E2, the Astrobee starts off in the Kibo capsule, but instead is given an single declarative knowledge statement \inquote{the harmony capsule is connected to the columbus capsule}, and then the same command.
\end{enumerate}

Similar to \cite{rosser2020a}, the language model was trained on a small corpus of 15 annotated sentences and referential expressions consisting of 16 unique words composing a total of 81 phrases.
The set of semantic concepts $\Gamma$ consisted of 34 different symbols.
All experiments were executed on a computer running Ubuntu 20.04 with 96 Intel Xeon CPUs 2.40GHz and 62.5 GiB of RAM and all processes consume a single thread.

\section{Results}

%For 2 hypothetical rollouts it took a total of x seconds

%In this time it will check synthesis, if unrealizable do rollouts, pause timer for input from user, restart timer once input is received and end upon acceptance of true statement by user

%Time to call the parser including overhead

%Note the reason that the first dialogue time is roughly double that of the second is because it is taking account for 2 synthesis attempts, the first that is triggered by the action symbol arrival, and the second by the first hypothetical world.

%%%%%%%%

%When the specification is realizable the state machine produced is dictated by these state transitions no matter the initial states or combination of declarative knowledge and queries used to get there.

\begin{figure*}[!hbtp]
    \centering
        %\subfloat[]{\includegraphics[width=0.3\textwidth]{film_strip/astrobee_scene00001d.png}\label{fig:file_strip_a}}
    %\hfil
        \subfloat[]{\includegraphics[width=0.495\linewidth]{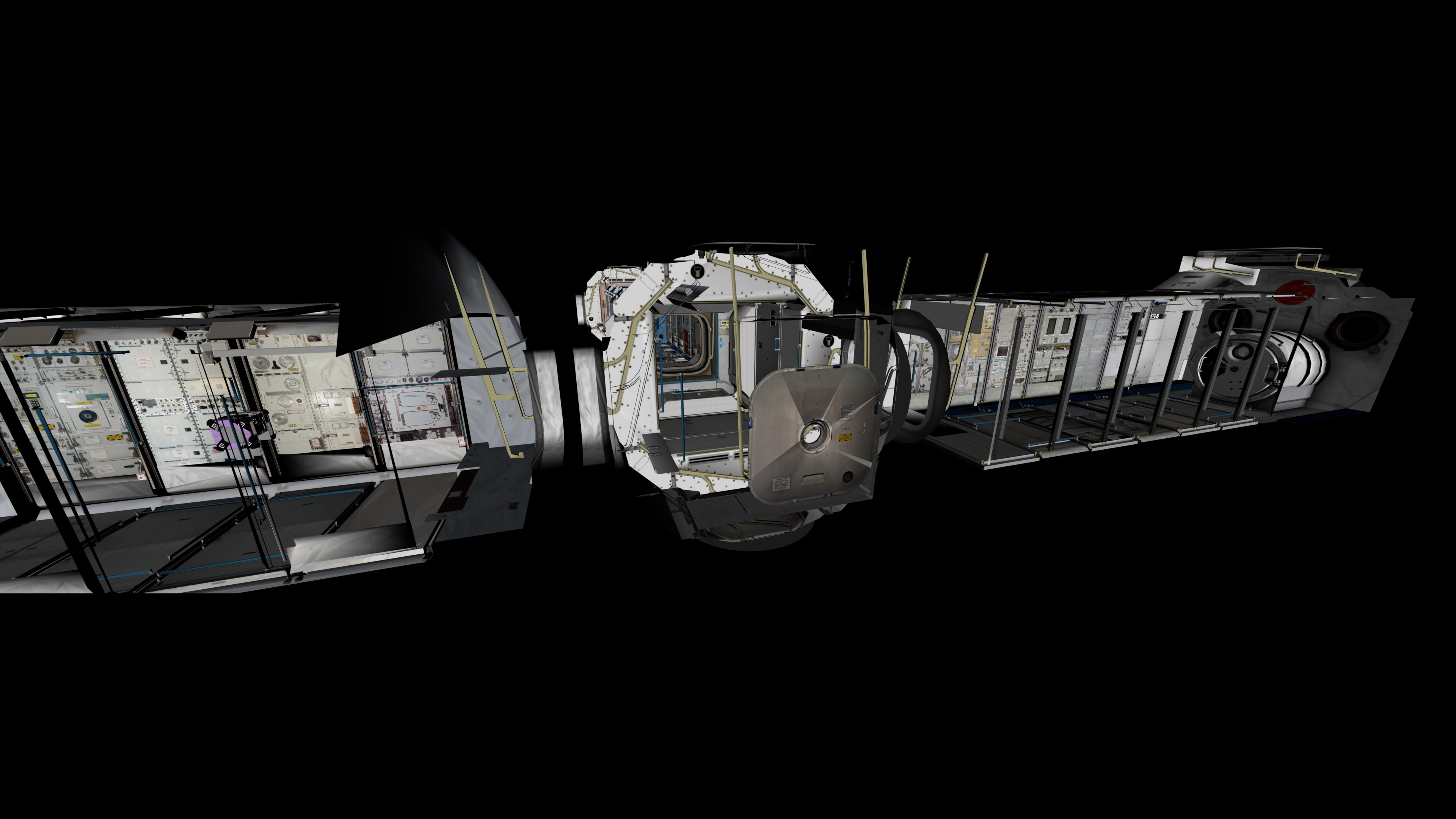}\label{fig:file_strip_a}}
    \hfil
        \subfloat[]{\includegraphics[width=0.495\linewidth]{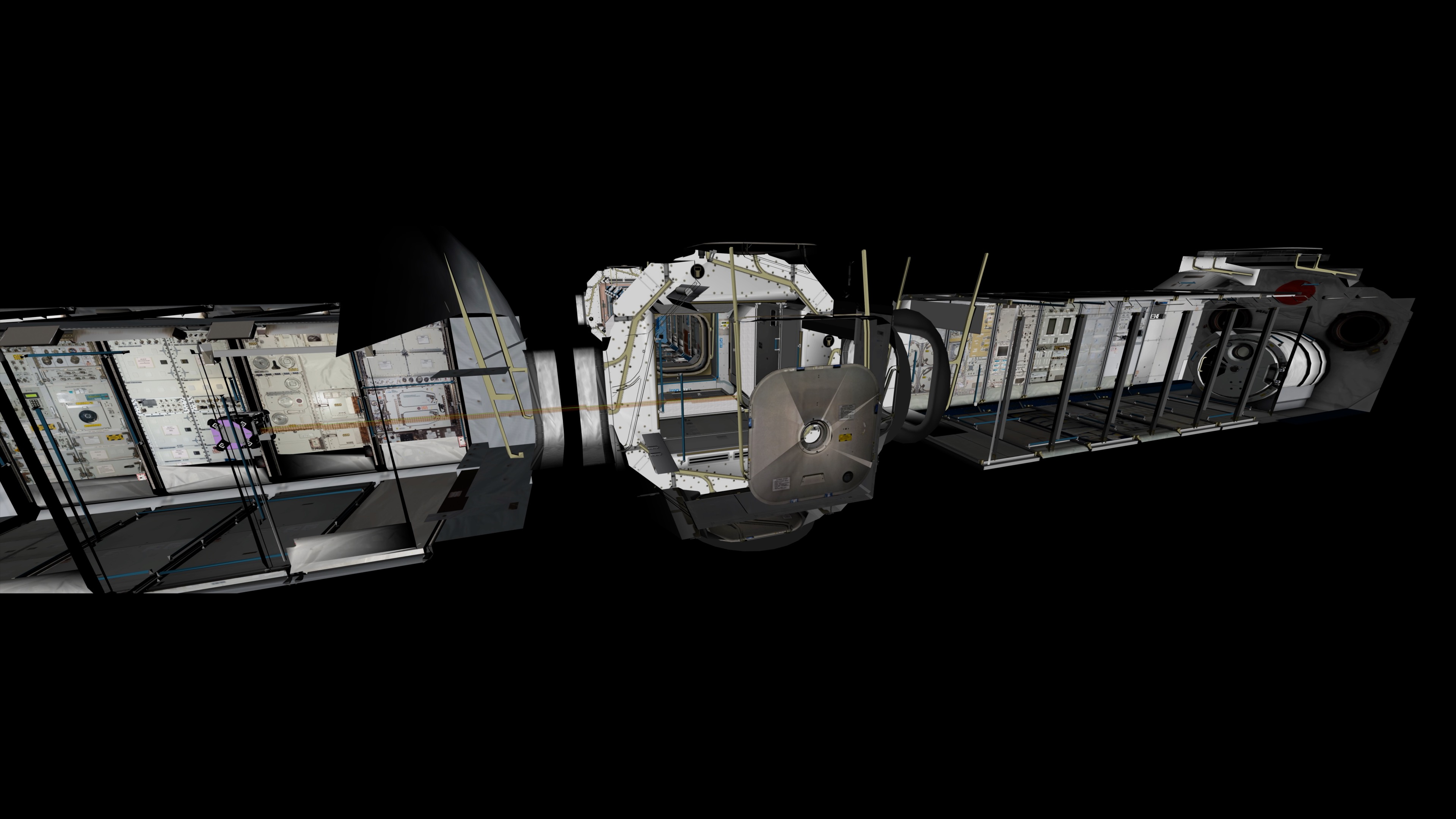}\label{fig:file_strip_b}}
        %\subfloat[]{\includegraphics[width=0.3\linewidth]{film_strip/astrobee_scene01251d.png}\label{fig:file_strip_d}}

        \subfloat[]{\includegraphics[width=0.495\textwidth]{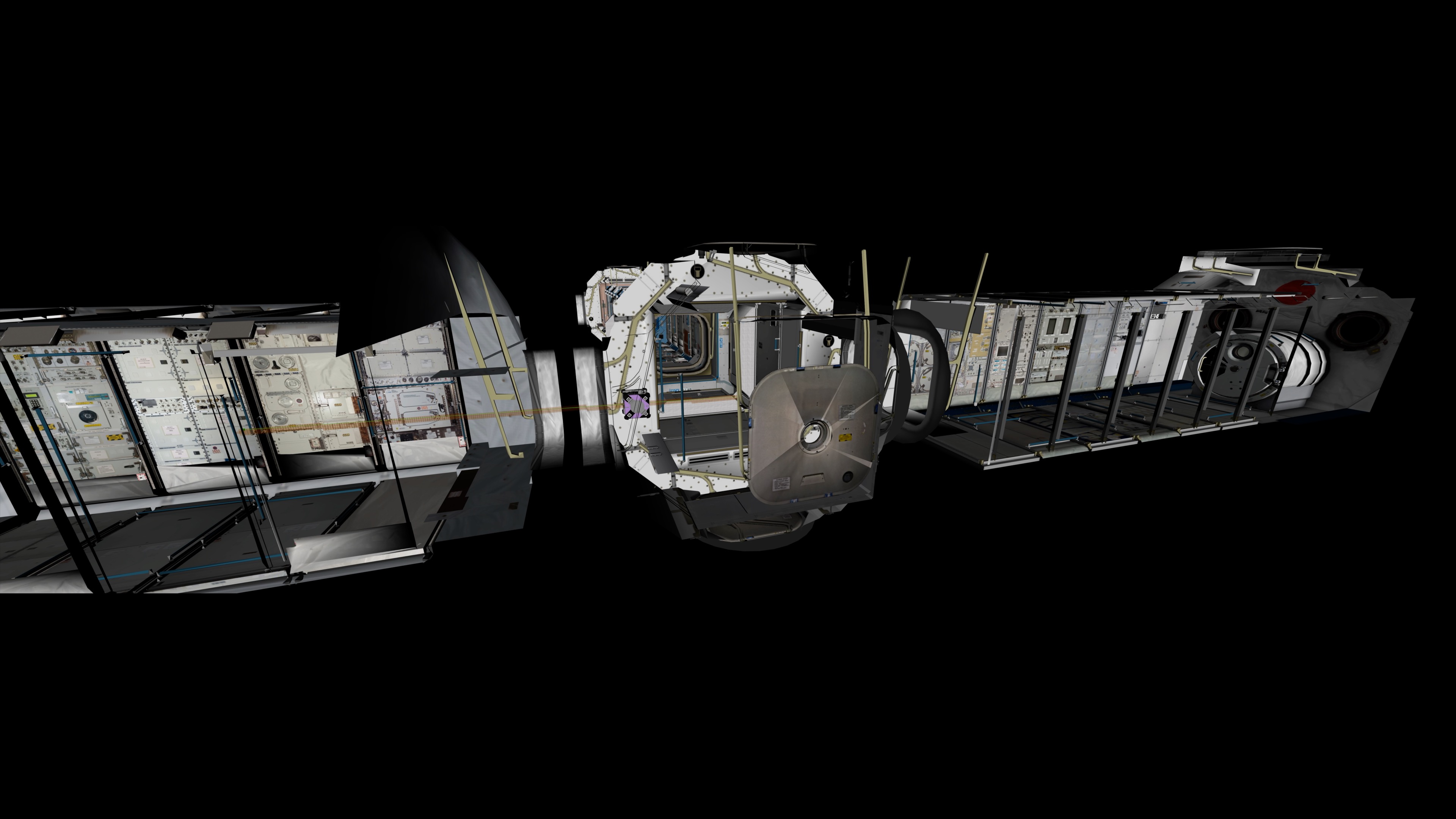}\label{fig:file_strip_c}}
    \hfil
        \subfloat[]{\includegraphics[width=0.495\linewidth]{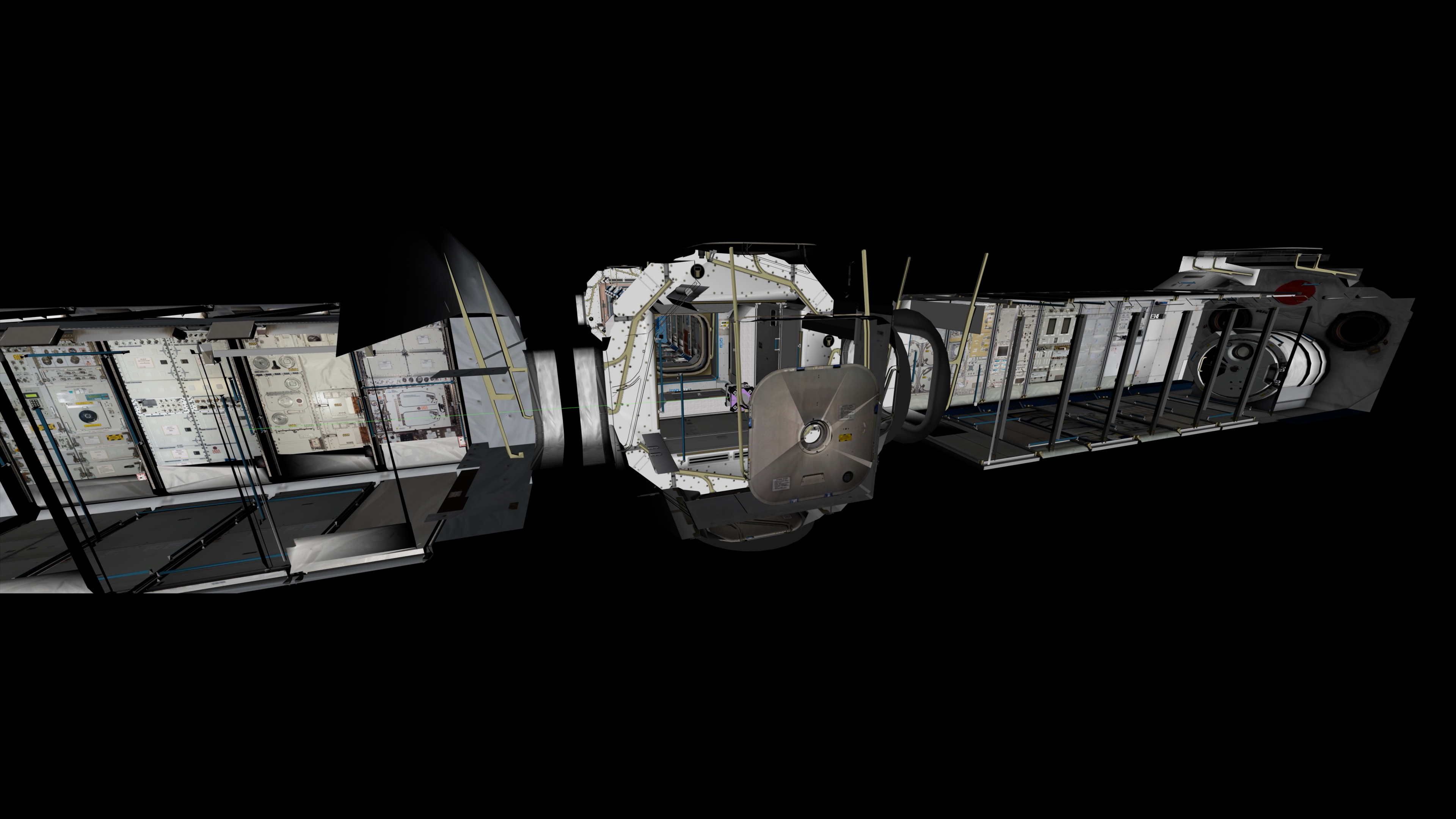}\label{fig:file_strip_d}}

        \subfloat[]{\includegraphics[width=0.495\textwidth]{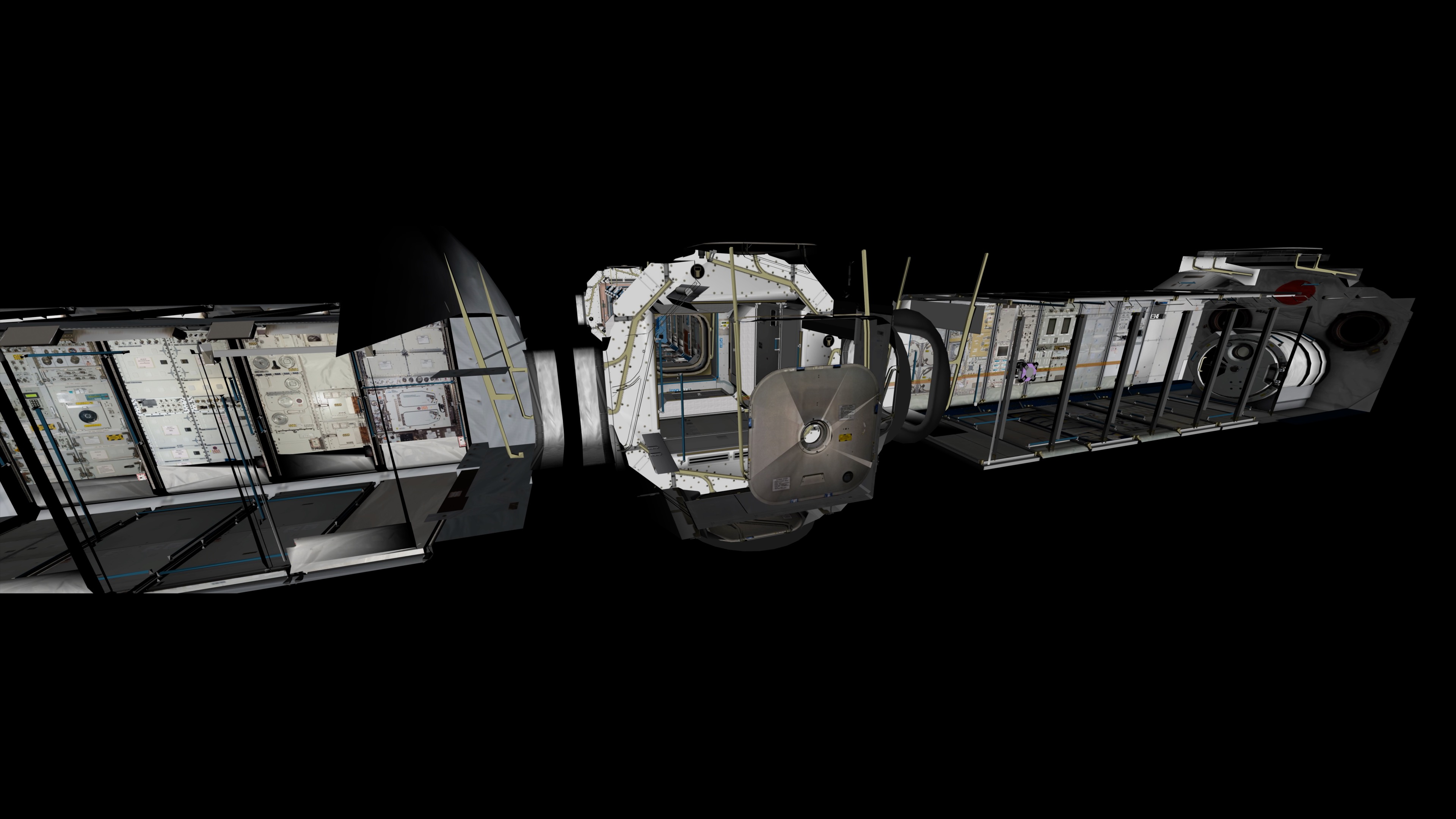}\label{fig:file_strip_e}}
    %\hfil
        %\subfloat[]{\includegraphics[width=0.3\linewidth]{film_strip/astrobee_scene07001d.png}\label{fig:file_strip_h}}
    \hfil
        \subfloat[]{\includegraphics[width=0.495\linewidth]{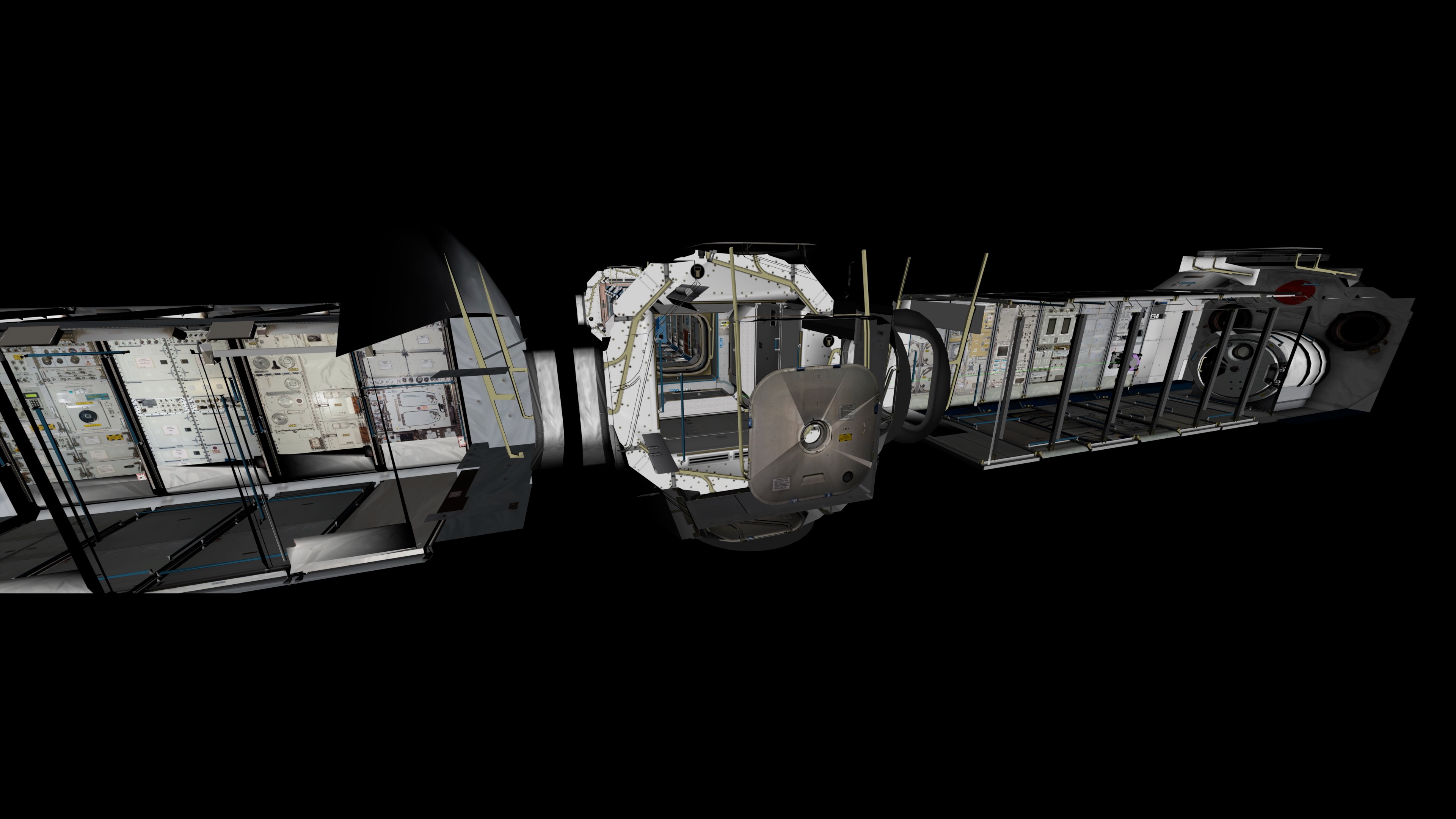}\label{fig:file_strip_f}}
    \caption{ \normalsize \protect\subref{fig:file_strip_a} The Astrobee is at rest at the centroid of the Columbus capsule awaiting a command from the controller, currently in the state \texttt{in\_columbus} \texttt{go\_columbus}
    \protect\subref{fig:file_strip_b} The Astrobee is given begins to move to the centroid of the Harmony capsule. \texttt{in\_columbus} \texttt{go\_harmony}
    \protect\subref{fig:file_strip_c} The Astrobee enters Harmony.
    \protect\subref{fig:file_strip_d} The Astrobee reaches the centroid of the Harmony capsule. \texttt{in\_harmony} \texttt{go\_harmony}, followed by being commanded to go the Kibo capsule via the controller \texttt{in\_harmony} \texttt{go\_kibo}
    \protect\subref{fig:file_strip_e} The Astrobee arrives at Kibo capsule and heads towards the centroid.
    \protect\subref{fig:file_strip_f} The final destination is reached by the Astrobee, thereby completing the command \texttt{in\_kibo} \texttt{go\_kibo}.
        \label{fig:film_strip} }
    \end{figure*}

\begin{figure*}[!hbtp]
    \centering
        %\subfloat[]{\includegraphics[width=0.3\textwidth]{film_strip/astrobee_scene00001d.png}\label{fig:file_strip_a}}
    %\hfil
        \subfloat[]{\includegraphics[width=0.495\linewidth]{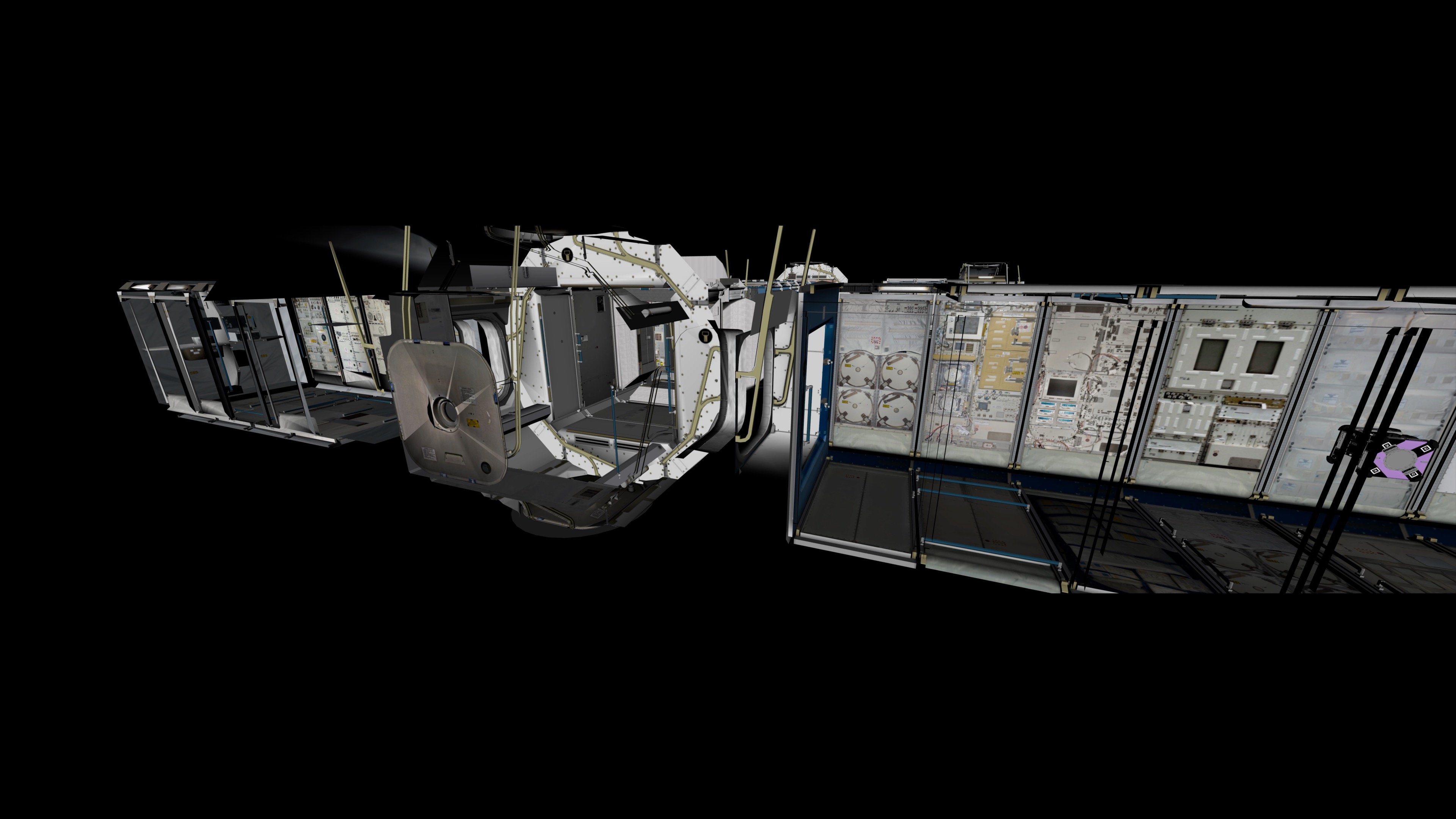}\label{fig:file_strip_2a}}
    \hfil
        \subfloat[]{\includegraphics[width=0.495\linewidth]{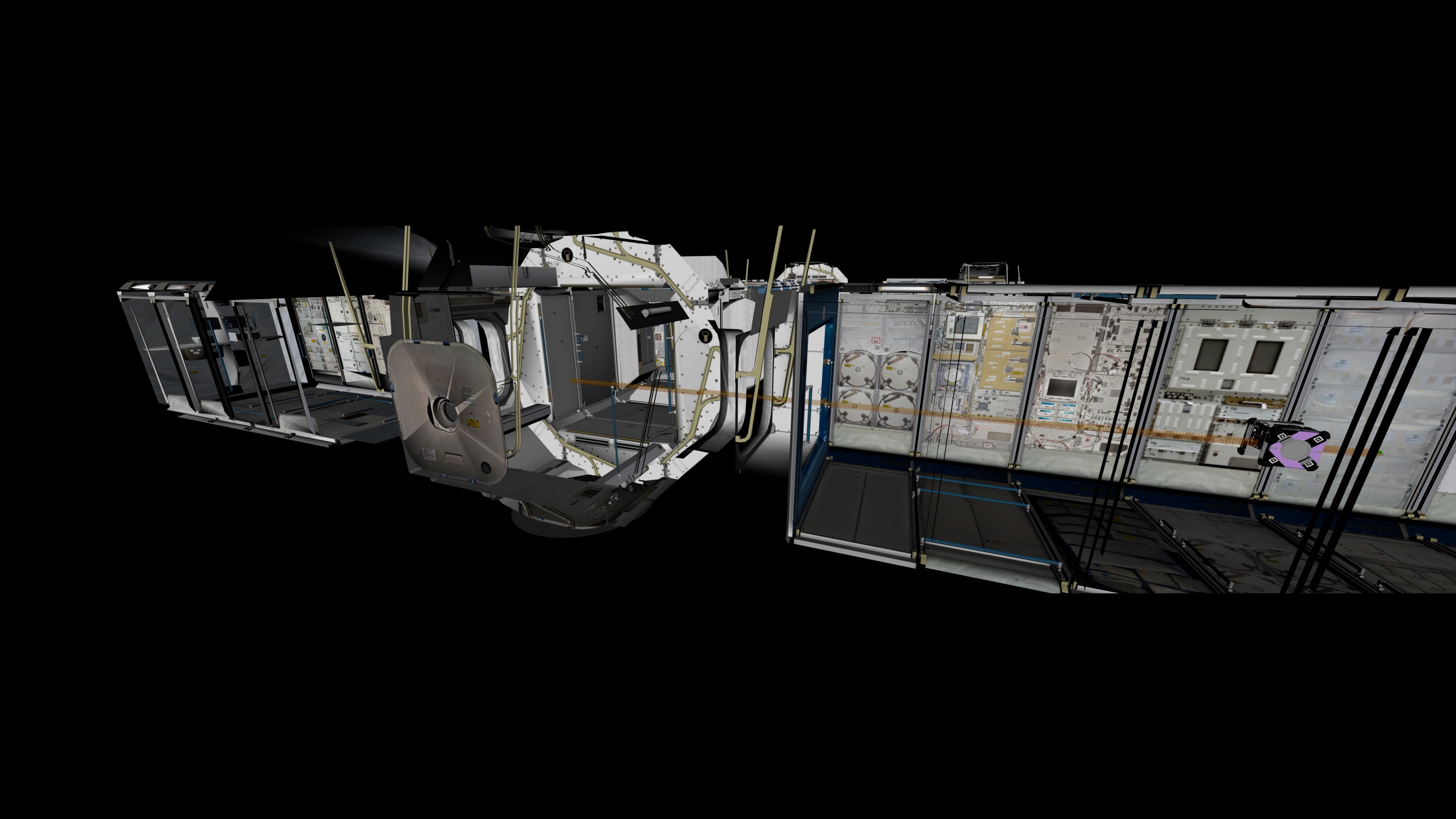}\label{fig:file_strip_2b}}
        %\subfloat[]{\includegraphics[width=0.3\linewidth]{film_strip/astrobee_scene01251d.png}\label{fig:file_strip_d}}
        
        \subfloat[]{\includegraphics[width=0.495\textwidth]{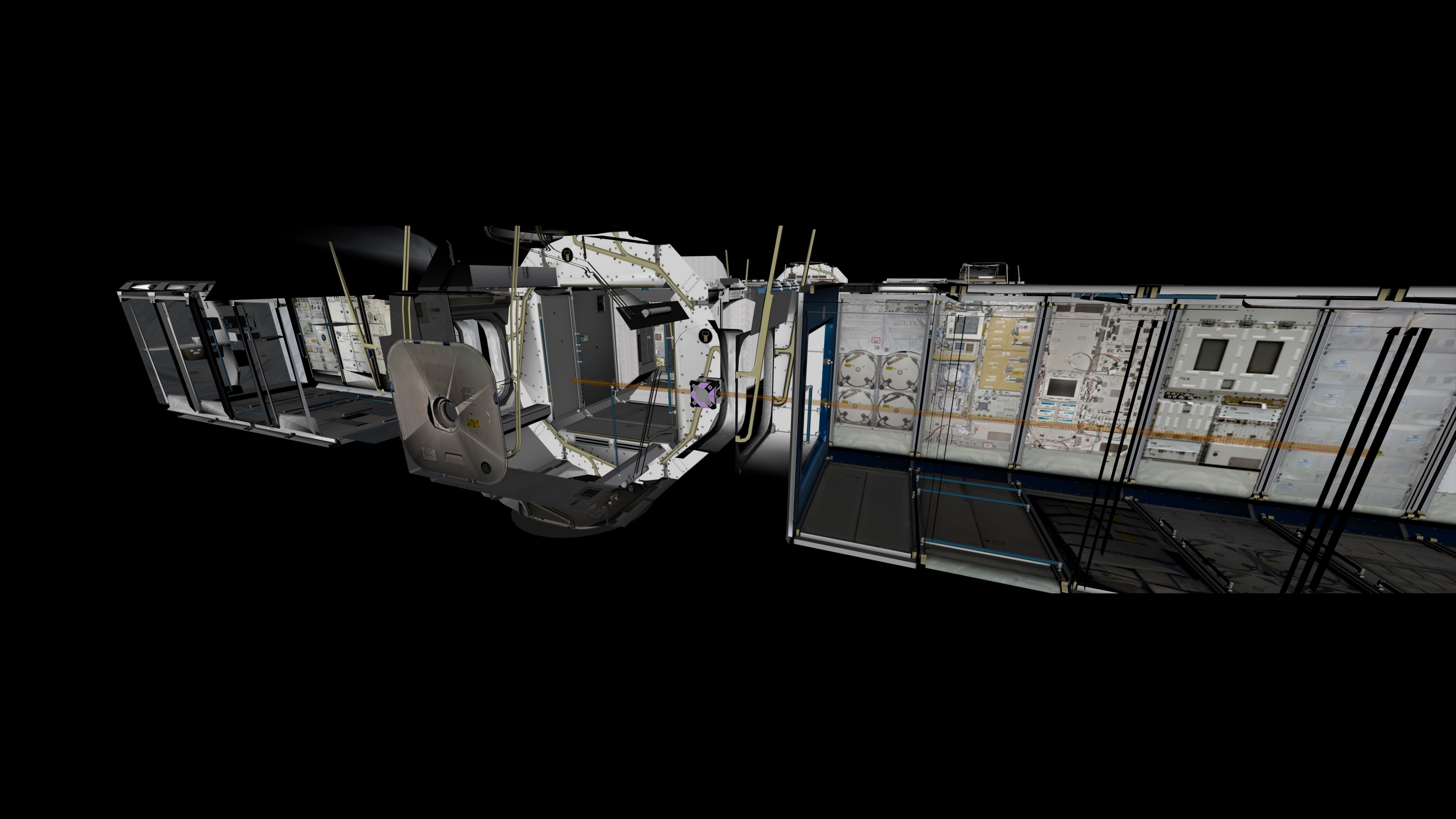}\label{fig:file_strip_2c}}
    \hfil
        \subfloat[]{\includegraphics[width=0.495\linewidth]{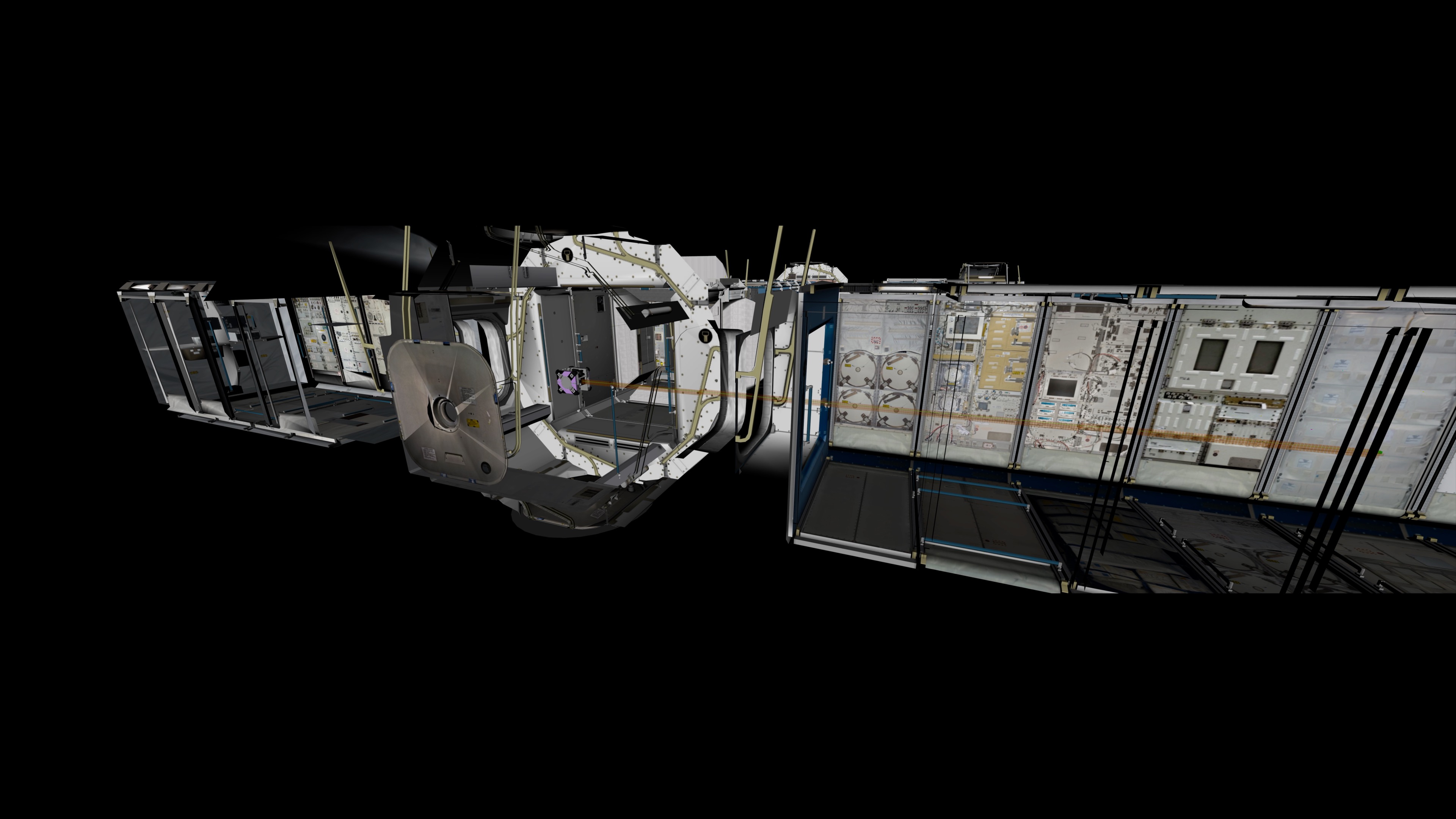}\label{fig:file_strip_2d}}

        \subfloat[]{\includegraphics[width=0.495\textwidth]{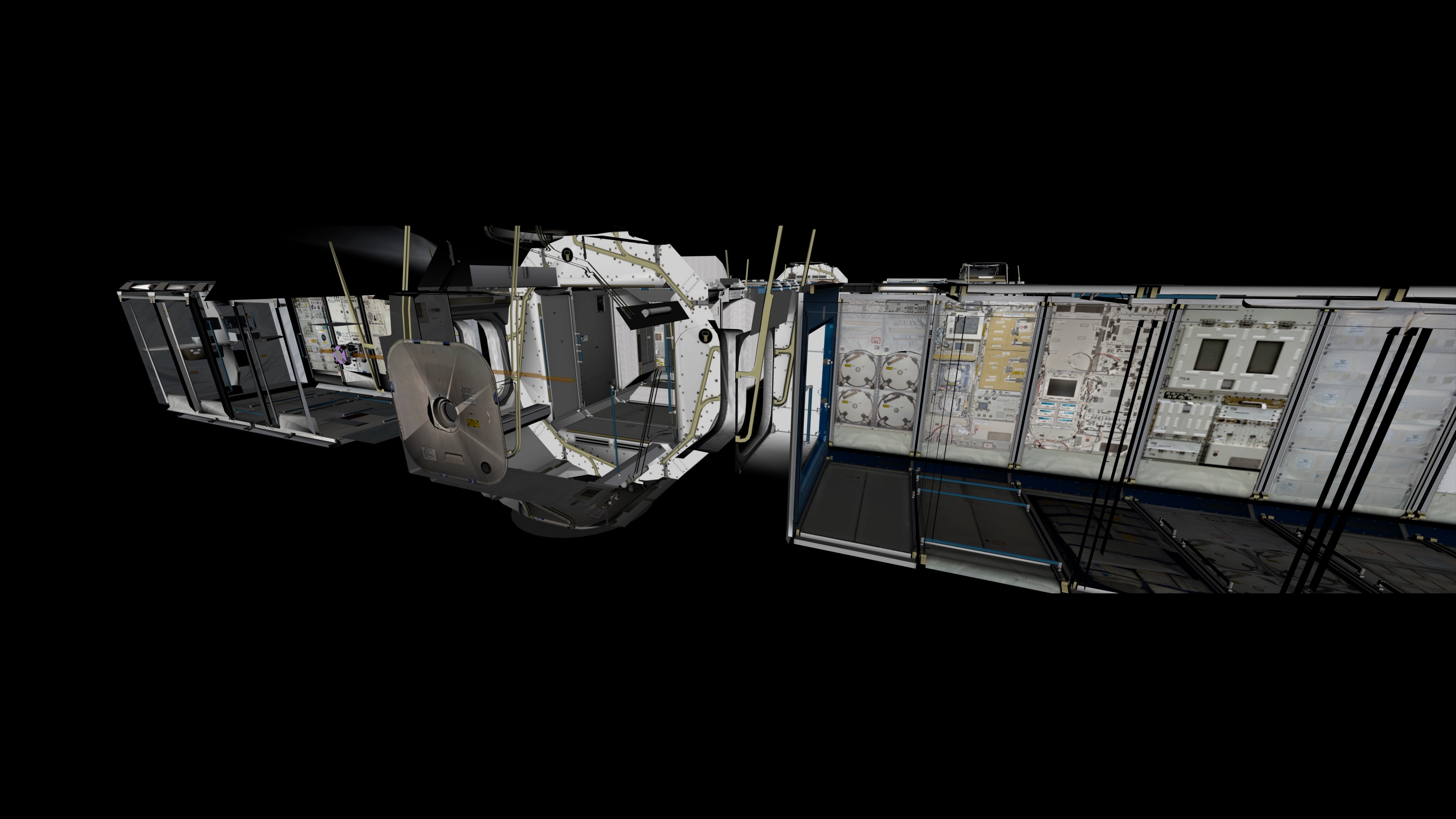}\label{fig:file_strip_2e}}
    %\hfil
        %\subfloat[]{\includegraphics[width=0.3\linewidth]{film_strip/astrobee_scene07001d.png}\label{fig:file_strip_h}}
    \hfil
        \subfloat[]{\includegraphics[width=0.495\linewidth]{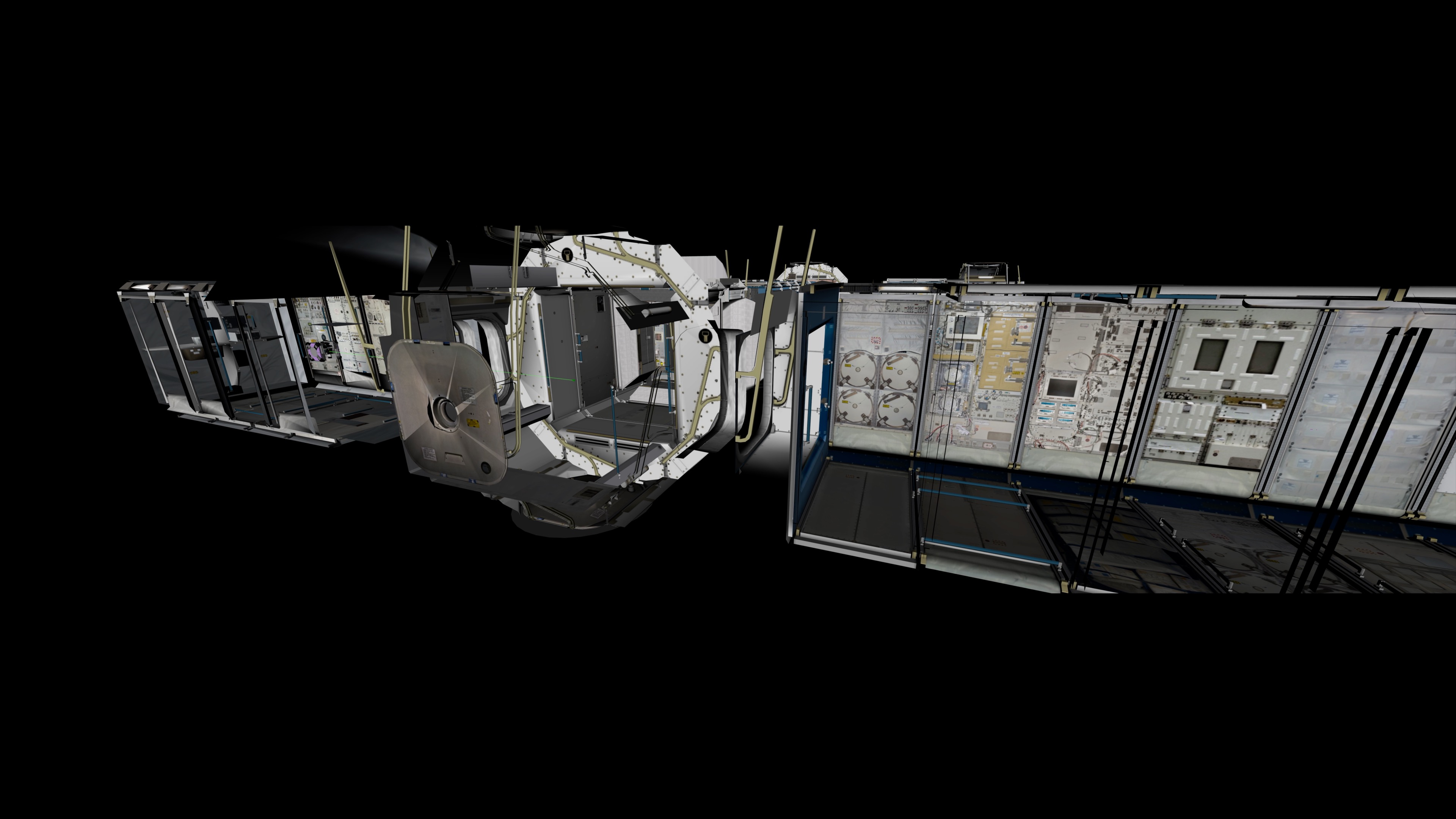}\label{fig:file_strip_2f}}
    \caption{ \normalsize \protect\subref{fig:file_strip_2a} The Astrobee is at the centroid of the Kibo module \texttt{in\_kibo} \texttt{go\_kibo}.
    \protect\subref{fig:file_strip_2b} The Astrobee heads to the Harmony capsule \texttt{in\_kibo} \texttt{go\_harmony}.
    \protect\subref{fig:file_strip_2c} The Astrobee arrives at Harmony and heads to the centroid.
    \protect\subref{fig:file_strip_2d} The Astrobee in the Harmony capsule \texttt{in\_harmony} \texttt{go\_harmony} followed by \texttt{in\_harmony} \texttt{go\_columbus}.
    \protect\subref{fig:file_strip_2e} The Astrobee enters the Columbus capsule and continues on to the centroid.
    \protect\subref{fig:file_strip_2f} The final destination is reached by the Astrobee, thereby completing the command \texttt{in\_columbus} \texttt{go\_columbus}.
        \label{fig:film_strip2} }
    \end{figure*}

\subsection{Experiment 1}
The interaction for the first scenario where two declarative knowledge statements ``the Kibo capsule is connected to the Harmony capsule'' and ``the Harmony capsule is connected to the Columbus capsule'' is followed by the instruction ``go to the Columbus capsule'' is illustrated below in Table \ref{tab1}.
The time taken to ground each of the declarative knowledge statements is 0.200 seconds, and the time to ground the instruction is 0.104 seconds.
The number of phrases of the instruction is roughly half the length of a declarative knowledge statement, thus it is expected that grounding the command takes approximately half the time as the declarative.
The time to update the internal logical representation of the world, write the specification, and synthesize the controller is 1.408 seconds.
In this scenario, the robot is able to immediately begin navigating to the goal because the two declarative knowledge statements provide enough logic to synthesize a controller. 

\begin{table}[h]
\begin{tabular}{|p{0.75cm}|p{1.0cm}|p{5cm}|}
\hline
Order & Speaker & Statement \\
\hline
1 & Human & ``the Kibo capsule is connected to the Harmony capsule'' \\
2 & Human & ``the Harmony capsule is connected to the Columbus capsule'' \\
3 & Human & ``go to the Kibo capsule'' \\
\hline
\end{tabular}
\caption{The interaction between a human and free-flying robot in the scenario illustrated in Figure \ref{fig:film_strip} given the declarative knowledge statements ``the Kibo capsule is connected to the Harmony capsule'' and ``the Harmony capsule is connected to the Columbus capsule''  is followed by the instruction ``go to the Kibo capsule''.}
\label{tab1}
\end{table}

\begin{figure}[hbtp]
    \centering
    \includegraphics[width=0.8\columnwidth]{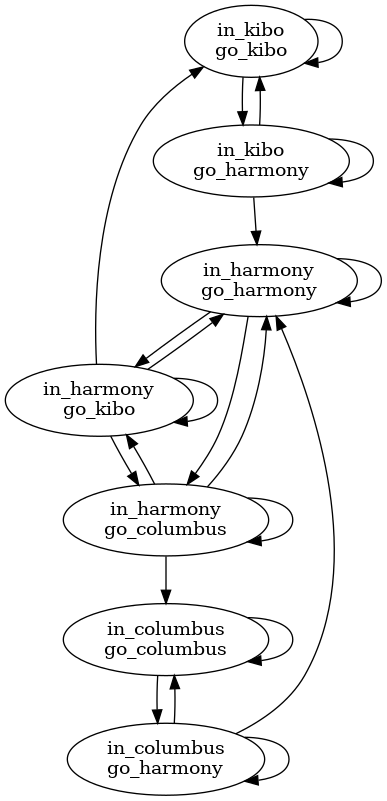}
    \caption{State Transitions permitted by synthesized controller}
    \label{fig:states}
\end{figure}

The state transitions permitted within the synthesized controller generated from the specification interpreted from language is illustrated in Figure \ref{fig:states}.
\texttt{in\_<capsule>} represents the input propositions as dictated by the location of the robot being true, where \texttt{go\_<capsule>} represents the output propositions true for that state indicative of the move action of the robot being taken in that state.
Prior to any linguistic interactions the robot is unaware of any connection between the three capsules only that they exist in the world model.
Therefore the only states that would exist are those where the robot is completing a move action for the capsule it is in i.e. $\left(\texttt{in\_kibo}, \texttt{go\_kibo}\right)$.
After the first declarative knowledge statement is provided the graph would then include edges allowing traversability between the Kibo and Harmony capsules i.e. $\left(\texttt{in\_kibo}, \texttt{go\_harmony}\right) \Rightarrow \left(\texttt{in\_harmony}, \texttt{go\_harmony}\right)$.
Similarly, after the second declarative knowledge statement is provided the graph would then include edges allowing traversability between the Columbus and Harmony capsules i.e. $\left(\texttt{in\_columbus}, \texttt{go\_harmony}\right) \Rightarrow \\ \left(\texttt{in\_harmony}, \\ \texttt{go\_harmony}\right)$.
The simulated Astrobee successfully completing the task to traverse from the Columbus capsule to the Kibo capsule as guided by the state machine in Figure \ref{fig:states} is displayed in Figure \ref{fig:film_strip}.
This experiment, which reproduces the approach described in \cite{rosser2020a}, provides a baseline to demonstrate that given two declarative knowledge statements, the world model can be filled in by the human to enable the robot to traverse from its current capsule to the goal.

\subsection{Experiment 2}
The interaction for the second scenario where the declarative knowledge statements ``the Kibo capsule is connected to the Harmony capsule'' and instruction ``go to the Columbus capsule'' is illustrated below in Table \ref{tab2}.  
The amount of time needed needed to ground statement 1 is 0.216 seconds and the amount of time needed to ground statement 2 is 0.112 seconds.
The time needed to update the internal logic representation from the command, write the specification, check synthesis, generate the first hypothetical world, and check its synthesis is 5.496 seconds.
The time needed to generate the natural language query is 1.432 seconds.
The human response of yes or no is simply analyzed by the NLP as a true or false statement requiring no NLU and passed along to the State Controller in Figure~\ref{fig:system_dia}, therefore a time is not reported.
The time needed to generate the second hypothetical world, write the specification, check its synthesis is 1.712 seconds.
The time needed to generate the second natural language query is 1.488 seconds.
All responses to queries were gated to exclude the buffering time of the human thinking and typing into the chat interface.

\begin{table}[h]
\begin{tabular}{|p{0.75cm}|p{1.0cm}|p{5cm}|}
\hline
Order & Speaker & Statement \\
\hline
1 & Human & ``the Kibo capsule is connected to the Harmony capsule'' \\
2 & Human & ``go to the Columbus capsule'' \\
3 & Robot & ``is the Kibo capsule connected to the Columbus capsule?'' \\
4 & Human & ``no'' \\
5 & Robot & ``is the Harmony capsule connected to the Columbus capsule?'' \\
6 & Human & ``yes'' \\
\hline
\end{tabular}
\caption{The dialogue between a human and free-flying robot in the scenario illustrated in Figure \ref{fig:film_strip2} given the declarative knowledge statement ``the Kibo capsule is connected to the Harmony capsule'' and instruction ``go to the Columbus capsule''.}
\label{tab2}
\end{table}
The state machine generated is identical to that of experiment 1, but instead derived from the querying process instead of from pure declarative knowledge statements.
The successful execution of experiment 2 is illustrated in Figure~\ref{fig:film_strip2} with the Astrobee traveling from the Kibo capsule to the Columbus capsule by way of the Harmony capsule.

\subsection{Experiment 3}
The interaction for the third scenario where the declarative knowledge statements ``the Harmony capsule is connected to the Columbus capsule'' and instruction ``go to the Columbus capsule'' is illustrated below in Table \ref{tab3}.  
The time required to ground statement 1 is 0.208 seconds and statement 2 0.104 seconds.
The time needed to update the internal logic representation from the command, write the specification, check synthesis, generate the first hypothetical world, and check its synthesis is 3.296 seconds.
The time required to generate the query is 1.344 seconds.
The main difference between the experiment 3 and experiment 2 is that only one query is needed to repair the specification, therefore it is an overall shorter total interaction.

\begin{table}[h]
\begin{tabular}{|p{0.75cm}|p{1.0cm}|p{5cm}|}
\hline
Order & Speaker & Statement \\
\hline
1 & Human & ``the Harmony capsule is connected to the Columbus capsule'' \\
2 & Human & ``go to the Columbus capsule'' \\
3 & Robot & ``is the Harmony capsule connected to the Kibo capsule?'' \\
4 & Human & ``yes'' \\
\hline
\end{tabular}
\caption{The dialogue between a human and free-flying robot in the scenario given the declarative knowledge statement ``the Harmony capsule is connected to the Columbus capsule'' and instruction ``go to the Columbus capsule''.}
\label{tab3}
\end{table}
The resultant behavior of the Astrobee in experiment 3 is identical of that in experiment 2 therefore a figure is not provided.
From experiment 3 it is clear that no matter which piece of declarative knowledge is provided first: \inquote{the Kibo capsule is connected to the Harmony capsule} or \inquote{the Harmony capsule is connected to the Columbus capsule} the system is able to generate targeted questions to arrive to the same state machine from Figure \ref{fig:states}, consistent with that produced in experiment 1.

\section{Conclusions}
In this work we demonstrate that a human and a robot can work towards consolidating mental models when faced with missing information through a combination of declarative knowledge statements and targeted questioning.
By using LTL driven verifiable controller synthesis, in combination with a set of semantic concepts from our language model, the robot can generate a set of hypothetical worlds to guide question generation to resolve ambiguities about its world model.
This was demonstrated in simulation with an Astrobee robot knowing the existence of 3 capsules in the ISS, but unaware of the connection between said capsules.
After receiving a single declarative knowledge statement followed by a commanded navigation task, we show that this pipeline enables the Astrobee to ultimately generate a state machine that completes the task consistent with the robot receiving purely declarative knowledge statements.
In future work we plan to increase the scale and complexity of the language model with a larger corpus to enable more freeform interactions.
Additionally, we hope to expand the systems capabilities beyond navigation tasks to include the ability to complete inspection tasks in which the robot must understand the behavior of an onboard astronaut to complete the inspection.
Finally, we hope to expand the pipeline to handle more than a single logical jump when reasoning over hypothetical worlds.
For example, one could imagine instead of providing a declarative knowledge statement to bridge the first logical gap, instead hypothesizing two additions to the world model to repair the misunderstanding with just targeted questions.

\bibliographystyle{IEEEtran}
\bibliography{ltliii.bib}

\begin{thebibliography}{10}
\providecommand{\url}[1]{#1}
\csname url@rmstyle\endcsname
\providecommand{\newblock}{\relax}
\providecommand{\bibinfo}[2]{#2}
\providecommand\BIBentrySTDinterwordspacing{\spaceskip=0pt\relax}
\providecommand\BIBentryALTinterwordstretchfactor{4}
\providecommand\BIBentryALTinterwordspacing{\spaceskip=\fontdimen2\font plus
\BIBentryALTinterwordstretchfactor\fontdimen3\font minus
  \fontdimen4\font\relax}
\providecommand\BIBforeignlanguage[2]{{%
\expandafter\ifx\csname l@#1\endcsname\relax
\typeout{** WARNING: IEEEtran.bst: No hyphenation pattern has been}%
\typeout{** loaded for the language `#1'. Using the pattern for}%
\typeout{** the default language instead.}%
\else
\language=\csname l@#1\endcsname
\fi
#2}}

\bibitem{Goldberg2002}
S.~Goldberg, M.~Maimone, and L.~Matthies, ``Stereo vision and rover navigation
  software for planetary exploration,'' in \emph{Proceedings, IEEE Aerospace
  Conference}, vol.~5, 2002, pp. 5--5.

\bibitem{Fong}
\BIBentryALTinterwordspacing
T.~Fong, M.~Bualat, L.~Edwards, L.~Flueckiger, C.~Kunz, S.~Lee, E.~Park, V.~To,
  H.~Utz, N.~Ackner, N.~Armstrong-Crews, and J.~Gannon, \emph{Human-Robot Site
  Survey and Sampling for Space Exploration}. [Online]. Available:
  \url{https://arc.aiaa.org/doi/abs/10.2514/6.2006-7425}
\BIBentrySTDinterwordspacing

\bibitem{Pedersen2006}
L.~Pedersen, W.~J. Clancey, M.~Sierhuis, N.~Muscettola, D.~E. Smith, D.~Lees,
  K.~Rajan, S.~Ramakrishnan, P.~Tompkins, A.~Vera, \emph{et~al.}, ``Field
  demonstration of surface human-robotic exploration activity.'' in \emph{AAAI
  Spring Symposium: To Boldly Go Where No Human-Robot Team Has Gone Before},
  2006, p. 114.

\bibitem{trevino2000first}
R.~C. Trevino, J.~J. Kosmo, A.~Ross, and N.~A. Cabrol, ``First astronaut-rover
  interaction field test,'' SAE Technical Paper, Tech. Rep., 2000.

\bibitem{Sierhuis2005}
M.~Sierhuis, W.~J. Clancey, R.~L. Alena, D.~Berrios, S.~Buckingham, J.~Dowding,
  J.~Graham, R.~V. Hoof, C.~Kaskiris, S.~Rupert, and K.~S. Tyree, ``Nasa’s
  mobile agents architecture: A multiagent workflow and communication system
  for planetary exploration,'' in \emph{8th International Symposium on
  Artificial Intelligence, Robotics and Automation in Space}, 2005.

\bibitem{Medina2011}
A.~Medina, C.~Pradalier, G.~Paar, A.~Merlo, S.~Ferraris, L.~Mollinedo,
  P.~Colmenarejo, and F.~Didot, ``A servicing rover for planetary outpost
  assembly,'' \emph{Astra}, 2011.

\bibitem{Diftler2007}
M.~A. {Diftler}, R.~O. {Ambrose}, W.~J. {Bluethmann}, F.~J. {Delgado},
  E.~{Herrera}, J.~J. {Kosmo}, B.~A. {Janoiko}, B.~H. {Wilcox}, J.~A.
  {Townsend}, J.~B. {Matthews}, T.~W. {Fong}, M.~G. {Bualat}, S.~Y. {Lee},
  J.~T. {Dorsey}, and W.~R. {Doggett}, ``{Crew/Robot Coordinated Planetary EVA
  Operations at a Lunar Base Analog Site},'' in \emph{38th Annual Lunar and
  Planetary Science Conference}, ser. Lunar and Planetary Science Conference,
  Mar. 2007, p. 1937.

\bibitem{Fong2006}
T.~Fong, J.~Scholtz, J.~A. Shah, L.~Fluckiger, C.~Kunz, D.~Lees, J.~Schreiner,
  M.~Siegel, L.~M. Hiatt, I.~Nourbakhsh, R.~Simmons, B.~Antonishek,
  M.~Bugajska, R.~Ambrose, R.~Burridge, A.~Schultz, and J.~G. Trafton, ``A
  preliminary study of peer-to-peer human-robot interaction,'' in \emph{2006
  IEEE International Conference on Systems, Man and Cybernetics}, vol.~4, 2006,
  pp. 3198--3203.

\bibitem{Howard2022}
T.~Howard, E.~Stump, J.~Fink, J.~Arkin, R.~Paul, D.~Park, S.~Roy, D.~Barber,
  R.~Bendell, K.~Schmeckpeper, J.~Tian, J.~Oh, M.~Wigness, L.~Quang,
  B.~Rothrock, J.~Nash, M.~Walter, F.~Jentsch, and N.~Roy, ``An intelligence
  architecture for grounded language communication with field robots,''
  \emph{Field Robotics}, vol.~2, no.~1, pp. 468--512, mar 2022.

\bibitem{howard14a}
T.~Howard, S.~Tellex, and N.~Roy, ``A natural language planner interface for
  mobile manipulators,'' in \emph{2014 IEEE International Conference on
  Robotics and Automation}.\hskip 1em plus 0.5em minus 0.4em\relax IEEE, May
  2014, pp. 6652--6659.

\bibitem{paul18a}
R.~Paul, J.~Arkin, D.~Aksaray, N.~Roy, and T.~M. Howard, ``Efficient grounding
  of abstract spatial concepts for natural language interaction with robot
  platforms,'' \emph{International Journal of Robotics Research}, vol.~37,
  no.~10, pp. 1269--1299, June 2018.

\bibitem{tellex2011a}
S.~Tellex, T.~Kollar, S.~Dickerson, M.~R. Walter, A.~Banerjee, S.~Teller, and
  N.~Roy, ``Approaching the symbol grounding problem with probabilistic
  graphical models,'' \emph{AI Mag.}, vol.~32, pp. 64--76, 2011.

\bibitem{clarke1996a}
E.~Clarke and J.~Wing, ``Formal methods: State of the art and future
  directions,'' \emph{ACM Computing Surveys}, vol.~28, 12 1996.

\bibitem{Manna1991TheTL}
Z.~Manna and A.~Pnueli, ``The temporal logic of reactive and concurrent
  systems,'' in \emph{Springer New York}, 1991.

\bibitem{Pnueli1977TheTL}
A.~Pnueli, ``The temporal logic of programs,'' \emph{18th Annual Symposium on
  Foundations of Computer Science (sfcs 1977)}, pp. 46--57, 1977.

\bibitem{smith16a}
T.~Smith, J.~Barlow, M.~Bualat, T.~Fong, C.~Provencher, H.~Sanchez, E.~Smith,
  and T.~A. Team, ``Astrobee: A new platform for free-flying robotics research
  on the international space station,'' in \emph{Proc. Int. Symp. on AI,
  Robotics, and Automation in Space (i-SAIRAS)}, June 2016.

\bibitem{Gao2016}
Y.~Gao, Ed., \emph{\BIBforeignlanguage{en}{Contemporary planetary
  robotics}}.\hskip 1em plus 0.5em minus 0.4em\relax Weinheim, Germany:
  Wiley-VCH Verlag, Aug. 2016.

\bibitem{cardoso2021verification}
R.~C. Cardoso, M.~Farrell, G.~Kourtis, M.~Webster, L.~A. Dennis, C.~Dixon,
  M.~Fisher, and A.~Lisitsa, ``Verification for space robotics,'' \emph{Space
  Robotics and Autonomous Systems: Technologies, Advances and Applications},
  vol. 131, p. 377, 2021.

\bibitem{kress2021formalizing}
H.~Kress-Gazit, K.~Eder, G.~Hoffman, H.~Admoni, B.~Argall, R.~Ehlers,
  C.~Heckman, N.~Jansen, R.~Knepper, J.~K{\v{r}}et{\'\i}nsk{\`y},
  \emph{et~al.}, ``Formalizing and guaranteeing human-robot interaction,''
  \emph{Communications of the ACM}, vol.~64, no.~9, pp. 78--84, 2021.

\bibitem{kress2018a}
\BIBentryALTinterwordspacing
H.~Kress-Gazit, M.~Lahijanian, and V.~Raman, ``Synthesis for robots: Guarantees
  and feedback for robot behavior,'' \emph{Annual Review of Control, Robotics,
  and Autonomous Systems}, vol.~1, no.~1, pp. 211--236, 2018. [Online].
  Available: \url{https://doi.org/10.1146/annurev-control-060117-104838}
\BIBentrySTDinterwordspacing

\bibitem{Clarke2018}
E.~Clarke and R.~Bloem, Eds., \emph{\BIBforeignlanguage{English}{Handbook of
  Model Checking}}.\hskip 1em plus 0.5em minus 0.4em\relax Springer, 2018.

\bibitem{Hoare1969}
\BIBentryALTinterwordspacing
C.~A.~R. Hoare, ``An axiomatic basis for computer programming,'' \emph{Commun.
  ACM}, vol.~12, no.~10, p. 576–580, oct 1969. [Online]. Available:
  \url{https://doi.org/10.1145/363235.363259}
\BIBentrySTDinterwordspacing

\bibitem{LEUCKER2009293}
\BIBentryALTinterwordspacing
M.~Leucker and C.~Schallhart, ``A brief account of runtime verification,''
  \emph{The Journal of Logic and Algebraic Programming}, vol.~78, no.~5, pp.
  293--303, 2009, the 1st Workshop on Formal Languages and Analysis of
  Contract-Oriented Software (FLACOS’07). [Online]. Available:
  \url{https://www.sciencedirect.com/science/article/pii/S1567832608000775}
\BIBentrySTDinterwordspacing

\bibitem{AraizaIllan2015}
D.~Araiza-Illan, D.~Western, A.~Pipe, and K.~Eder, ``Coverage-driven
  verification ---,'' in \emph{Hardware and Software: Verification and
  Testing}, N.~Piterman, Ed.\hskip 1em plus 0.5em minus 0.4em\relax Cham:
  Springer International Publishing, 2015, pp. 69--84.

\bibitem{Salem2015}
M.~Salem, G.~Lakatos, F.~Amirabdollahian, and K.~Dautenhahn, ``Would you trust
  a (faulty) robot?: Effects of error, task type and personality on human-robot
  cooperation and trust,'' vol. 2015, 03 2015.

\bibitem{kress2009a}
H.~{Kress-Gazit}, G.~E. {Fainekos}, and G.~J. {Pappas}, ``Temporal-logic-based
  reactive mission and motion planning,'' \emph{IEEE Transactions on Robotics},
  vol.~25, no.~6, pp. 1370--1381, 2009.

\bibitem{pnueli1989a}
\BIBentryALTinterwordspacing
A.~Pnueli and R.~Rosner, ``On the synthesis of a reactive module.''\hskip 1em
  plus 0.5em minus 0.4em\relax New York, NY, USA: Association for Computing
  Machinery, 1989. [Online]. Available:
  \url{https://doi.org/10.1145/75277.75293}
\BIBentrySTDinterwordspacing

\bibitem{rosser2020a}
J.~Rosser, J.~Arkin, S.~Patki, and T.~M. Howard, ``Natural language interaction
  with synthesis based control for simulated free-flying robots,'' in
  \emph{International Symposium on Artificial Intelligence, Robotics, and
  Automation for Space (ISAIRAS)}, Oct. 2020.

\bibitem{BLOEM2012911}
\BIBentryALTinterwordspacing
R.~Bloem, B.~Jobstmann, N.~Piterman, A.~Pnueli, and Y.~Sa'ar, ``Synthesis of
  reactive(1) designs,'' \emph{Journal of Computer and System Sciences},
  vol.~78, no.~3, pp. 911 -- 938, 2012, in Commemoration of Amir Pnueli.
  [Online]. Available:
  \url{http://www.sciencedirect.com/science/article/pii/S0022000011000869}
\BIBentrySTDinterwordspacing

\bibitem{ehlers16a}
R.~Ehlers and V.~Raman, ``Slugs: Extensible gr(1) synthesis,'' in
  \emph{Computer Aided Verification}, S.~Chaudhuri and A.~Farzan, Eds.\hskip
  1em plus 0.5em minus 0.4em\relax Cham: Springer International Publishing,
  2016, pp. 333--339.

\bibitem{ehlers15a}
R.~{Ehlers}, R.~{Könighofer}, and R.~{Bloem}, ``Synthesizing cooperative
  reactive mission plans,'' in \emph{2015 IEEE/RSJ International Conference on
  Intelligent Robots and Systems (IROS)}, 2015, pp. 3478--3485.

\bibitem{kress2008a}
\BIBentryALTinterwordspacing
H.~Kress-Gazit, G.~E. Fainekos, and G.~J. Pappas, ``Translating structured
  english to robot controllers,'' \emph{Advanced Robotics}, vol.~22, no.~12,
  pp. 1343--1359, 2008. [Online]. Available:
  \url{https://doi.org/10.1163/156855308X344864}
\BIBentrySTDinterwordspacing

\bibitem{finucane2010a}
C.~{Finucane}, {Gangyuan Jing}, and H.~{Kress-Gazit}, ``Ltlmop: Experimenting
  with language, temporal logic and robot control,'' in \emph{2010 IEEE/RSJ
  International Conference on Intelligent Robots and Systems}, 2010, pp.
  1988--1993.

\bibitem{Lignos2014a}
\BIBentryALTinterwordspacing
C.~Lignos, V.~Raman, C.~Finucane, M.~Marcus, and H.~Kress-Gazit, ``Provably
  correct reactive control from natural language,'' \emph{Autonomous Robots},
  vol.~38, no.~1, pp. 89--105, Nov. 2014. [Online]. Available:
  \url{https://doi.org/10.1007/s10514-014-9418-8}
\BIBentrySTDinterwordspacing

\bibitem{boteanu16a}
A.~Boteanu, J.~Arkin, T.~M. Howard, and H.~Kress-Gazit, ``A model for
  verifiable grounding and execution of complex language instructions,'' in
  \emph{2016 IEEE/RSJ International Conference on Intelligent Robots and
  Systems}.\hskip 1em plus 0.5em minus 0.4em\relax IEEE, Oct. 2016, pp.
  2649--2654.

\bibitem{boteanu17a}
A.~Boteanu, J.~Arkin, S.~Patki, T.~M. Howard, and H.~Kress-Gazit,
  ``Robot-initiated specification repair through grounded language
  interaction,'' in \emph{AAAI Fall Symposium on Natural Communication for
  Human-Robot Collaboration}, Nov. 2017.

\bibitem{arkin2020}
\BIBentryALTinterwordspacing
J.~Arkin, D.~Park, S.~Roy, M.~R. Walter, N.~Roy, T.~M. Howard, and R.~Paul,
  ``Multimodal estimation and communication of latent semantic knowledge for
  robust execution of robot instructions,'' \emph{The International Journal of
  Robotics Research}, vol.~0, no.~0, pp. 1--26, 2020. [Online]. Available:
  \url{https://doi.org/10.1177/0278364920917755}
\BIBentrySTDinterwordspacing

\bibitem{Williams2013}
T.~Williams, R.~Cantrell, G.~Briggs, P.~Schermerhorn, and M.~Scheutz,
  ``Grounding natural language references to unvisited and hypothetical
  locations,'' 01 2013, pp. 947--953.

\bibitem{raman2013b}
V.~{Raman} and H.~{Kress-Gazit}, ``Explaining impossible high-level robot
  behaviors,'' \emph{IEEE Transactions on Robotics}, vol.~29, no.~1, pp.
  94--104, 2013.

\bibitem{Raman-RSS-13}
V.~Raman, C.~Lignos, C.~Finucane, K.~C.~T. Lee, M.~Marcus, and H.~Kress-Gazit,
  ``Sorry dave, i'm afraid i can't do that: Explaining unachievable robot tasks
  using natural language,'' in \emph{Proceedings of Robotics: Science and
  Systems}, Berlin, Germany, June 2013.

\end{thebibliography}

\acknowledgments
This work was supported by an Early Career Faculty grant from NASA's Space Technology Research Grants Program.

%%%%%%%%%%%%%%%%%%%%%%%%%%%%%%%%%%%%%%%%%%%%%%%%%%%%%%%%%%%%%%%%%%%%%%%%%%%%%%%%%%%%%%%%%%%%%%%%%%%%%%
%\bibliography{IEEEabr,MyBibFile}

%%%%%%%%%%%%%%%%%%%%%%%%%%%%%%%%%%%%%%%%%%%%%%%%%%%%%%%%%%%%%%%%%%%%%%%%%%%%%%%%%%%%%%%%%%%%%%%%%%%%%%
\thebiography
%% This biostyle allows you to insert your photo size 1in X 1.25in

\begin{biographywithpic}
{Joshua Rosser}{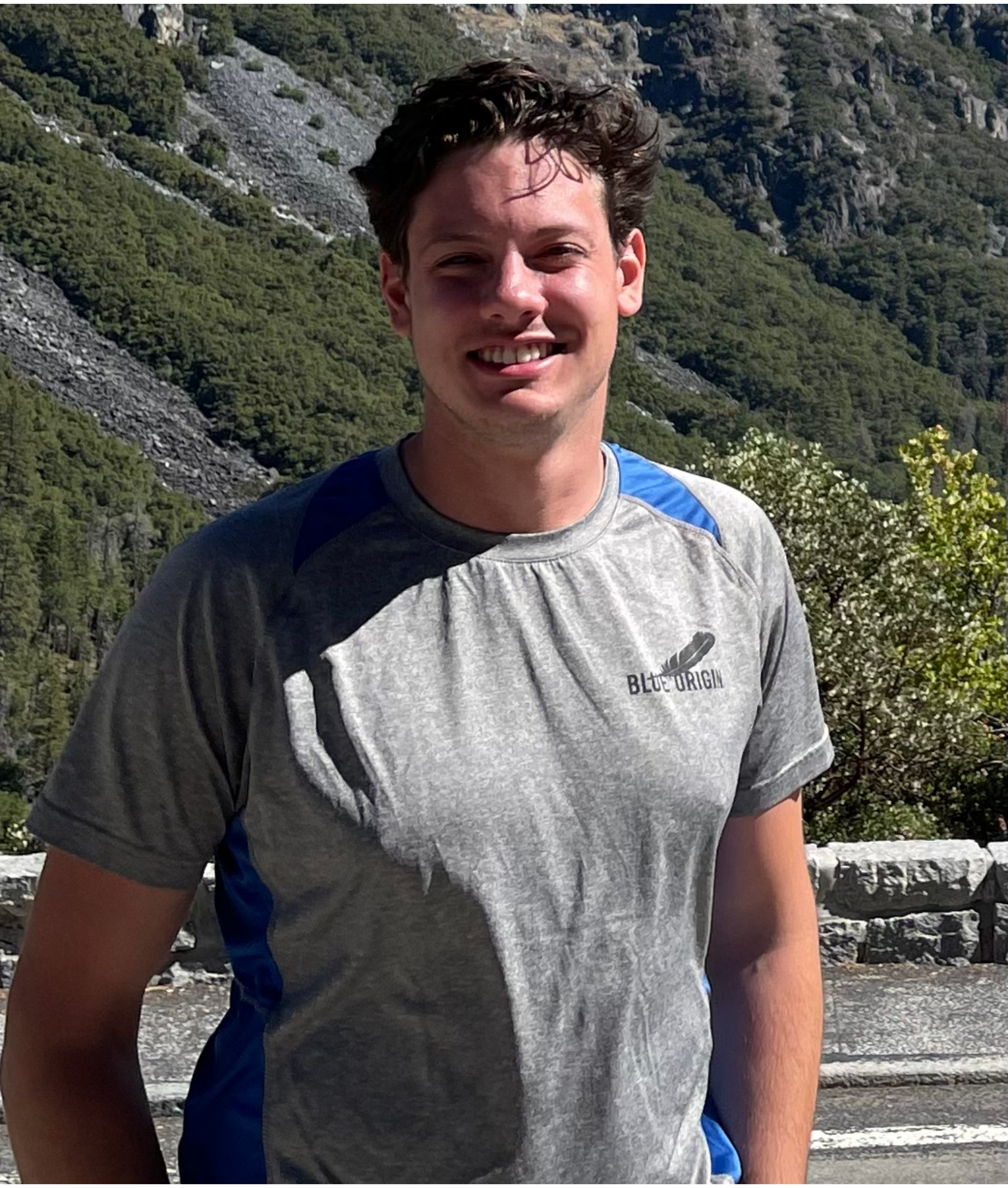}
is a current PhD student in Electrical and Computer Engineering at the University of Rochester. He has a B.S. in Physics and Astronomy also from the University of Rochester. Research interests are in motion planning in unstructured environments, HRI, task planning for field and space robotics.
\end{biographywithpic}

\begin{biographywithpic}
{Jacob Arkin}{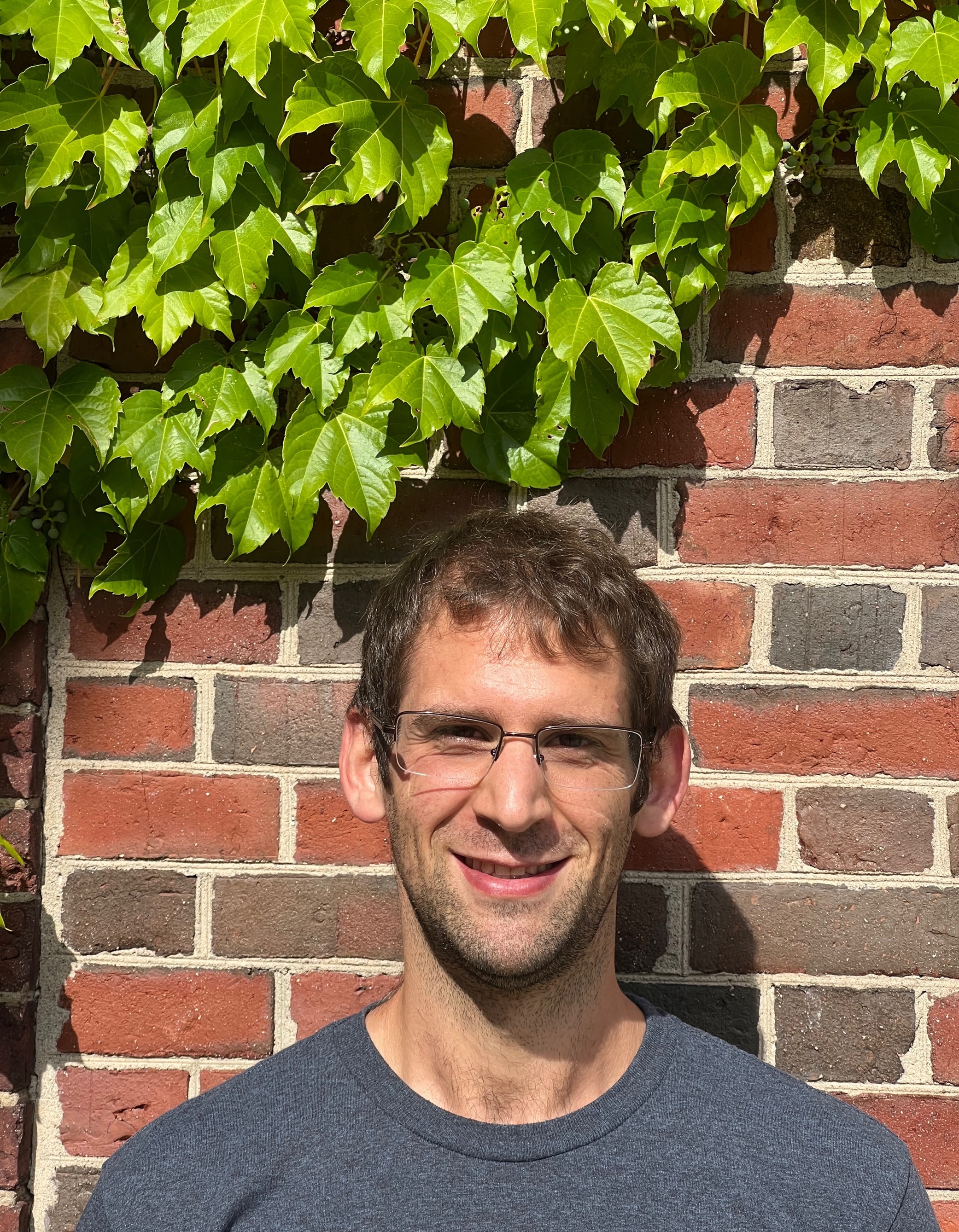}
is a Ph.D. student in the University of Rochester’s Department of Electrical and Computer Engineering, having received both a Bachelor’s and Master’s Degree of ECE from the University. His research primarily focuses on bidirectional language as a mode of communication and collaboration for human-robot teams.
\end{biographywithpic}

\begin{biographywithpic}
{Siddharth Patki}{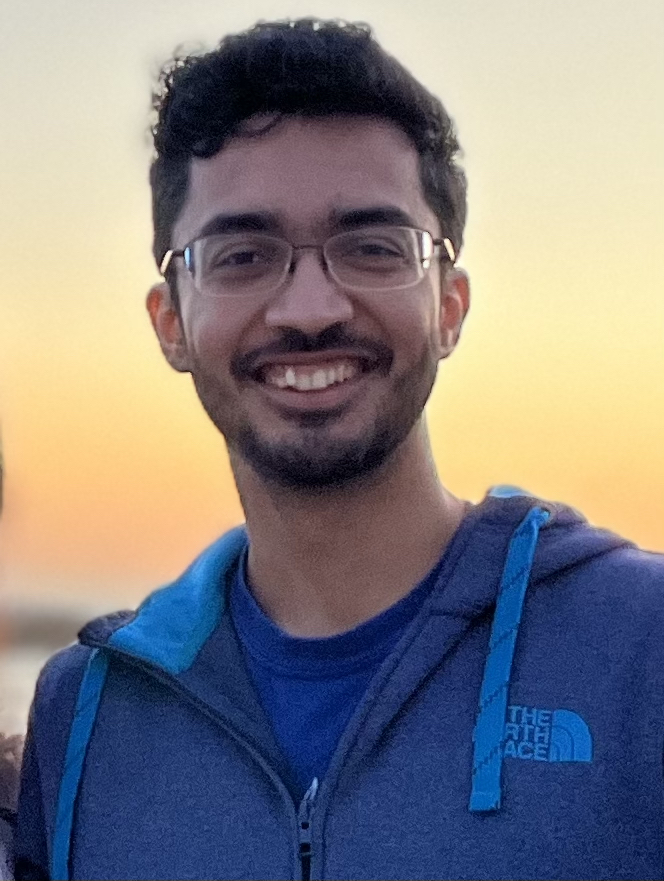}
is a PhD student in the Electrical and Computer Engineering Department at the University of Rochester. He received his B.Tech degree in Electronics Engineering from the University of Pune, India. His current research focuses on developing adaptive models of perception for enabling efficient natural language based interaction with collaborative robots in dynamic environments.
\end{biographywithpic}

\begin{biographywithpic}
{Thomas Howard}{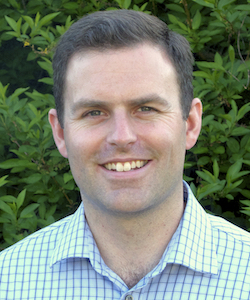}
is an Associate Professor of Electrical and Computer Engineering at the University of Rochester.  Previously, he was a Research Scientist and a Postdoctoral Associate at MIT's Computer Science and Artificial Intelligence Laboratory, a Research Technologist II at the Jet Propulsion Laboratory, and a Lecturer in Mechanical Engineering at Caltech.  He earned his Ph.D. in Robotics from Carnegie Mellon University and holds Bachelor of Science degrees in Mechanical Engineering and Electrical and Computer Engineering from the University of Rochester. 
\end{biographywithpic}

\end{document}